\documentclass[runningheads]{llncs}

\usepackage{eccv}

\usepackage{eccvabbrv}

\usepackage{graphicx}
\usepackage{booktabs}

\usepackage{subcaption}
\usepackage{amsmath,bm}
\usepackage[accsupp]{axessibility}  

\usepackage{hyperref}

\usepackage{orcidlink}

\newcommand{\modelname}{Arc2Face}

\begin{document}

\title{\modelname{}: A Foundation Model for ID-Consistent Human Faces}

\author{Foivos Paraperas Papantoniou\inst{1} \and
Alexandros Lattas\inst{1} \and
Stylianos Moschoglou\inst{1} \and
Jiankang Deng\inst{1} \and
Bernhard Kainz\inst{1,2} \and
Stefanos Zafeiriou\inst{1}
}

\authorrunning{F.~Paraperas Papantoniou et al.}

\institute{Imperial College London, UK
\and
FAU Erlangen–N\"urnberg, Germany
\\
\textcolor{magenta}{\url{https://arc2face.github.io/}}
}

\maketitle

\begin{abstract}
  This paper presents \modelname{}, an identity-conditioned face foundation model, which, given the ArcFace embedding of a person, can generate diverse photo-realistic images with an unparalleled degree of face similarity than existing models. Despite previous attempts to decode face recognition features into detailed images, we find that common high-resolution datasets (\eg FFHQ) lack sufficient identities to reconstruct any subject. To that end, we meticulously upsample a significant portion of the WebFace42M database, the largest public dataset for face recognition (FR). \modelname{} builds upon a pretrained Stable Diffusion model, yet adapts it to the task of ID-to-face generation, conditioned solely on ID vectors. Deviating from recent works that combine ID with text embeddings for zero-shot personalization of text-to-image models, we emphasize on the compactness of FR features, which can fully capture the essence of the human face, as opposed to hand-crafted prompts. Crucially, text-augmented models struggle to decouple identity and text, usually necessitating some description of the given face to achieve satisfactory similarity. \modelname{}, however, only needs the discriminative features of ArcFace to guide the generation, offering a robust prior for a plethora of tasks where ID consistency is of paramount importance. As an example, we train a FR model on synthetic images from our model and achieve superior performance to existing synthetic datasets. 
\end{abstract}

\begin{figure}[t]
    \centering
    \includegraphics[width=\textwidth]{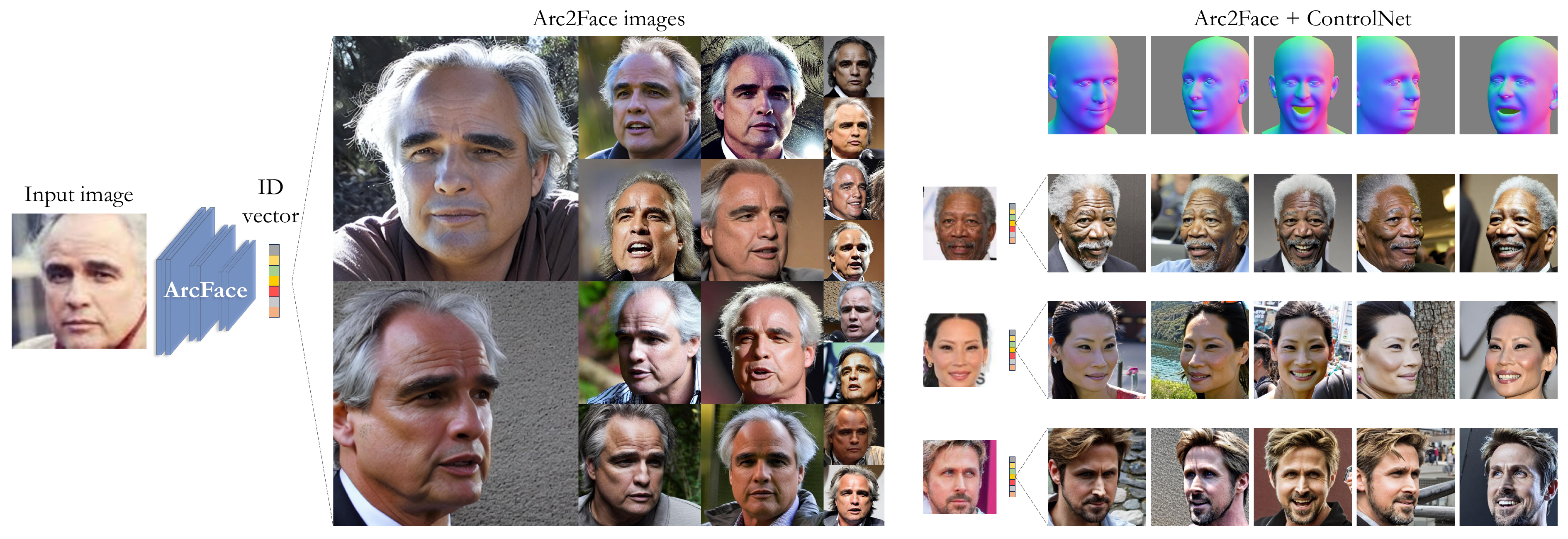}
    \caption{Given the ID-embedding from \cite{deng2019arcface}, \modelname{} can generate high-quality images of any subject with compelling similarity. Using popular extensions, such as ControlNet~\cite{zhang2023adding}, we can explicitly control facial attributes such as the pose or expression.}
    \label{fig:teaser}
    \vspace{-0.7cm}
\end{figure}

\section{Introduction}
\label{sec:intro}

Learning statistical priors from data to generate new facial images is a highly popular problem at the intersection of Computer Vision and Machine Learning. Arguably, the most well-known early method was the so-called Eigenfaces \cite{turk1991eigenfaces}, which applies the Karhunen-Loeve expansion, also known as Principal Component Analysis (PCA), to a set of facial images. The success of Eigenfaces with well-aligned frontal facial images captured under controlled conditions spurred a significant line of research into linear and nonlinear component analysis over 25 years ago. Probably the pinnacle of PCA applications was in 3D Morphable Models (3DMMs)~\cite{Blanz3DMM}, where it was used to learn a prior for facial textures represented in a UV map. However, when directly applied to 2D images, PCA had many shortcomings, including the requirement for images to be perfectly pixel-aligned. Furthermore, it could not describe hair (facial or not) and also struggled with non-linearities introduced by facial expressions and lighting conditions.

The recent evolution of deep neural networks (DNNs) has made it feasible to overcome many of the limitations inherent in simple linear statistical models, such as PCA, enabling the learning of statistical priors from large-scale facial data captured under varied recording conditions. Arguably, the most successful method, based on Generative Adversarial Networks (GANs) \cite{goodfellow2014generative}, is StyleGAN \cite{karras2019style} and its variants \cite{karras2020analyzing, karras2021alias}. StyleGAN learns a decoder governed by low-dimensional latent codes, which can be sampled to generate images. Moreover, StyleGAN can be used to fit images to its distributions (\ie, find the best latent code to generate a specific image) \cite{xia2022gan}, proving highly useful as a prior for many tasks, including 3D face reconstruction for facial texture generation \cite{ganfit_gecer}. Additionally, several studies have discussed how StyleGAN's latent spaces can be utilized for controllable manipulations (\ie, changing the expressions or the pose of facial images) \cite{shen2020interfacegan, harkonen2020ganspace, zhu2021low}. However, to date, identity drift remains a challenge in latent space manipulations using StyleGAN. 

One of the first problems to find a robust solution with the advent of deep learning is face recognition (FR), generally approached by identifying features that represent facial identities (also known as ID-embeddings). These features are so powerful that they are used in smartphones for face verification, leading to a vast research field. Arguably, the most widely used ID-embeddings are those produced by the ArcFace \cite{deng2019arcface} loss. ArcFace features have been utilized as perceptual features for 3D face reconstruction and generation. However, robustly controlling the identity within the StyleGAN framework 
remains challenging, mainly due to the necessity of large high-resolution datasets with significant intra-class variability.

In the past couple of years, another revolution concerning generative methods has occurred in deep learning. The emergence of the so-called Diffusion Models~\cite{diffusion2015} has demonstrated not only the possibility of modeling the distribution of images and sampling from it \cite{ddpm_Ho, song2021Score-SDE, dhariwal2021diffusion}, but also the feasibility of guiding the generation process with features that correlate images with textual descriptions, such as CLIP~\cite{radford2021learning, rombach2022high}. Recently, it has been shown that other features, \eg, ID-embeddings, can be incorporated to steer the generation process for faces, to not only adhere to textual descriptions but also to the facial characteristics of a person's image \cite{valevski2023face0, ye2023ip, wang2024instantid, chen2023dreamidentity, peng2023portraitbooth}.

A challenge arises when combining CLIP features with ID-embeddings because CLIP features contain ID-related information (otherwise, generating the image of a famous celebrity from a textual prompt would not be possible), as well as possess contradictory features to ID-embeddings (\eg, when requesting a photo of a ``Viking'' with the ID-embeddings of a Chinese person). Hence, although these models are useful for entertainment purposes and demonstrate the power of diffusion models, they are neither suitable for the controlled generation of facial images nor for sampling conditioned to the ID features of a specific subject.

In this paper, we meticulously study the problem of high-resolution facial image synthesis conditioned on ID-embeddings and propose a large-scale foundation model. Developing such a model poses a significant challenge due to the limited availability of high-quality facial image databases. Specifically:
\begin{itemize}
\renewcommand{\labelitemi}{$\bullet$}
    \item We show that smaller, single-image-per-person databases~\cite{karras2019style} are insufficient to train a robust foundation model, thus, we introduce a large dataset of high-resolution facial images with consistent identity and intra-class variability derived from WebFace42M.
    \item We adapt a large facial identity encoder trained \textsl{solely} on ID-embeddings, in contrast to text-based models where ID interferes with language.
    \item We introduce the first ID-conditioned face foundation model, which we make available to the public.
    \item We propose a robust benchmark for evaluating ID-conditioned models. Our experiments demonstrate superior performance compared to existing approaches.
\end{itemize}

\section{Related Work}
\label{sec:related}

\subsection{Generative Models for the Human Face}
\label{sec:related_generation}
The task of facial image generation has witnessed great success in recent years,
with the advent of style-based GANs \cite{karras2019style, karras2020analyzing, karras2021alias, mensah2023hybrid},
even for view-consistent synthesis\cite{chan2021pi, deng2022gram, gu2021stylenerf, niemeyer2021giraffe, chan2022efficient, schwarz2022voxgraf, an2023panohead}.
Despite early attempts to condition GANs on modalities such as text \cite{sauer2023stylegan, kang2023studiogan, kang2023scaling},
the recent Diffusion Models \cite{ddpm_Ho, song2021Score-SDE, dhariwal2021diffusion, yang2023diffusion} have shown tremendous progress. Trained on large-scale datasets, several prominent models, including 
DALLE-2~\cite{ramesh2022hierarchical}, Imagen~\cite{saharia2022photorealistic}, or Stable Diffusion~\cite{rombach2022high}, have emerged, marking a breakthrough in creative image generation. Later models have enabled an even higher level of detail \cite{podell2023sdxl}.
Nevertheless, dataset memorization remains an issue in such models \cite{somepalli2023diffusion}.
Moreover, recent methods have achieved even 3D human generation using 
general \cite{papantoniou2023relightify, kirschstein2023diffusionavatars}
and text-guided diffusion models \cite{huang2023humannorm, zhang2023dreamface}.

In terms of controlling the generation output, adding images alongside text has been proposed. For example, ILVR~\cite{choi2021ilvr} conditions sampling on a target image through iterative refinement, while SDEdit~\cite{meng2022sdedit} adds noise to the input before editing it according to the target description. DiffusionRig \cite{ding2023diffusionrig} enables explicit face editing, using a crude 3DMM \cite{feng2021learning} rendering to condition generation on pose, illumination, and expression. Perhaps the most notable approach, ControlNet~\cite{zhang2023adding}, first achieved accurate spatial control of text-to-image models,
with the addition of trainable copies of network parts.
In a different direction, universal guidance was introduced \cite{Bansal_2023_CVPR}
to avoid re-training the model during conditioning. While the combined use of text and spatial control can enable manipulation of an input photo to some extent, it cannot accurately generate one's identity under any conditions, which is addressed by subject-driven generative models.

\subsection{Subject-Specific Facial Generation}
\label{sec:related_one-shot}

When considering subject-conditioned generation,
face recognition (FR) models provide a powerful platform 
for identity features extraction from facial images, since they typically 
extract comprehensive facial embeddings in order to measure identity similarity
(\eg~\cite{deng2019arcface, boutros2022elasticface, kim2022adaface, deng2021variational}).
Therefore, the inversion of such models in a black-box setting has been shown capable of producing facial images from an identity embedding~\cite{razzhigaev2020black, vendrow2021realistic, mai2018reconstruction, yang2019neural}, 
even when using GAN and diffusion architectures for zero-shot generation~\cite{duong2020vec2face, truong2022vec2face, kansy2023controllable}. However, current inversion methods are either trained on low-resolution datasets~\cite{duong2020vec2face, truong2022vec2face} designed for face recognition or use high-quality but limited images~\cite{kansy2023controllable}, constraining their generalizability.

The most impressive results have recently been shown as extensions of Stable Diffusion~\cite{rombach2022high}.
The main paradigm emerged from Textual Inversion \cite{gal2022image} and especially DreamBooth \cite{ruiz2023dreambooth},
which fine-tunes a diffusion model on several images of a subject,
to learn a subject-specific class identifier reproducing that specific subject. Follow-up works reduced the optimization time, such as HyperDreamBooth \cite{ruiz2023hyperdreambooth}, which utilizes LoRA \cite{hu2021lora} and a 
hypernetwork to perform tuning on a single image. Similarly, E4T~\cite{gal2023encoder} and ProFusion~\cite{zhou2023enhancing} propose encoder-based approaches, while CustomDiffusion~\cite{kumari2023multi} optimizes only a subset of network's parameters. Moreover, Celeb-Basis \cite{yuan2023inserting} and StableIdentity \cite{wang2024stableidentity} learn an embedding basis from a set of celebrities that can be used to condition the text-based model.
In a generalized manner, Kosmos-G \cite{pan2023kosmos} introduced a multi-modal perception model that accepts various inputs, including facial images.

Closest to ours lies a series of recent works that condition Diffusion Models directly on facial features for tuning-free personalization. Various approaches, including FastComposer \cite{xiao2023fastcomposer}, PhotoVerse \cite{chen2023photoverse} and PhotoMaker \cite{li2023photomaker},
employ features from the CLIP \cite{radford2021learning} image encoder to represent the input subject. However, this representation is constrained by 
the CLIP's facial encoding abilities. 
Methods such as Face0 \cite{valevski2023face0}, DreamIdentity \cite{chen2023dreamidentity}, and PortraitBooth \cite{peng2023portraitbooth} condition the model additionally on FR embeddings for improved fidelity.
Similarly, IP-Adapter \cite{ye2023ip} uses a decoupled cross-attention mechanism to separate text from subject conditioning and has released, post-publication, \cite{ye2023ip-github} an impressive model based on ID features. InstantID \cite{wang2024instantid} extends \cite{ye2023ip-github} with an additional network for stronger ID guidance and facial landmarks conditioning. Finally, FaceStudio \cite{yan2023facestudio} 
learns combined CLIP and ID-embeddings,
achieving impressive stylizations.

A common concept in prior works is to use subject embeddings as an additional conditioning mechanism on top of text. Therefore, they exhibit a deficiency in identity retention due to the ambiguous relation of text description and fine identity characteristics. In contrast, in this work, we introduce a model solely conditioned on robust identity features,
achieving state-of-the-art control and identity retention, as shown in Fig.~\ref{fig:teaser}.

\begin{figure}[t]
    \vspace{-0.5cm}
    \centering
    \includegraphics[width=\textwidth]{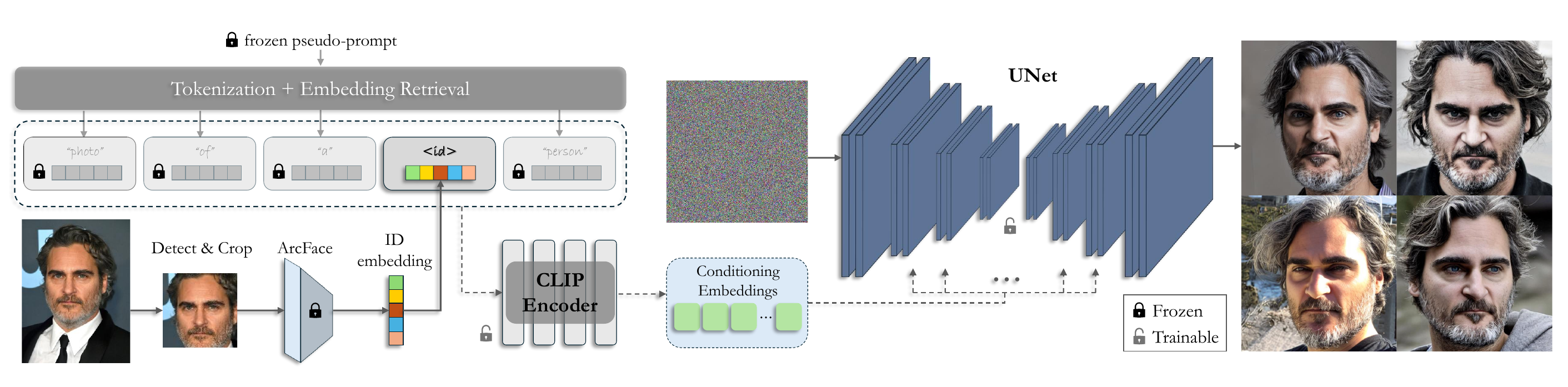}
    \caption{Overview of \modelname{}. We use a straightforward design to condition Stable Diffusion on ID features. The ArcFace embedding is processed by the text encoder using a frozen pseudo-prompt for compatibility, allowing projection into the CLIP latent space for cross-attention control. Both the encoder and UNet are optimized on a million-scale FR dataset~\cite{zhu2021webface260m} (after upsampling), followed by additional fine-tuning on high-quality datasets~\cite{karras2017progressive, karras2019style}, without any text annotations. \textbf{The resulting model exclusively adheres to ID-embeddings, disregarding its initial language guidance}.}
    \label{fig:method}
    \vspace{-0.5cm}
\end{figure}

\section{Method}
\label{sec:method}
Our objective is to develop a foundation model that accurately generates images of any subject independent of pose, expression, or contextual scene information. To that end, we employ the ArcFace \cite{deng2019arcface} network as our feature extractor, owing to its intrinsic ability to filter out such information. Our model extends a powerful Stable Diffusion \cite{rombach2022high} backbone, enabling efficient sampling of high-quality images. Specific details of our methodology are provided in the subsequent sections.

\subsection{Preliminaries}

\subsubsection{Face Recognition Features:}
The utilization of pre-trained FR networks to constrain face-related optimization or generation tasks has seen significant success in recent years. Typically, ID similarity is employed as an auxiliary loss, often by extracting features from multiple layers of these networks. In this work, we aim to utilize these networks as frozen feature extractors in order to condition a generative model of facial images. In particular, let $\phi$ denote the forward function of a pre-trained ArcFace~\cite{deng2019arcface} network. Given a cropped and aligned face image $x\in \mathbb{R}^{H\times W\times C}$, $\phi$ extracts a high-level embedding $\mathbf{w} = \phi(x) \in \mathbb{R}^{512}$, designed to separate the face from other subjects. We rely only on this vector to recover novel images of the subject, without access to any intermediate layers of~\cite{deng2019arcface}. As shown by Kansy \etal~\cite{kansy2023controllable}, it is possible to learn this mapping via a conditional diffusion model without 
any ID-specific loss functions. Similarly, we employ $\mathbf{w}$ to condition our model by projecting it onto the cross-attention layers of the generator, following the successful paradigm of Stable Diffusion.

\subsubsection{Latent Diffusion Models:}
Diffusion models~\cite{ddpm_Ho, song2021Score-SDE, dhariwal2021diffusion} employ a denoising mechanism to approximate the distribution of real images $x$. During training, images undergo distortion through the addition of Gaussian noise via a predetermined diffusion schedule at different timesteps $t$, given by $x_t = \sqrt{\bar{\alpha}_t} x_0 + (1-\bar{\alpha}t)\epsilon$. At the same time, a denoising autoencoder $\epsilon_{\theta}(x_t,t)$ is trained to recover the normally distributed noise $\epsilon\sim\mathcal{N}(\mathbf{0},\mathbf{I})$ by minimizing the prediction error :
\begin{equation}
    \mathcal{L} = \mathbb{E}_{x_t,t,\epsilon\sim \mathcal{N}(\mathbf{0},\mathbf{I})}\left[ || \epsilon - \epsilon_{\theta}(x_t, t) ||_2^2 \right]\label{eq:unet_loss}
\end{equation}
The Latent Diffusion (LD) model~\cite{rombach2022high} is a widely adopted architecture within the diffusion framework. Notably, a Variational Autoencoder (VAE), $\mathcal{E}$, is employed to compress images into a lower-dimensional latent space $z=\mathcal{E}(x)$ for efficient training. Moreover, LD introduces a universal conditioning mechanism, using a UNet~\cite{ronneberger2015u} as the backbone for $\epsilon_{\theta}$, with cross-attention layers to map auxiliary features $C$ to its intermediate layers for conditional noise prediction $\epsilon_{\theta}(z_t,t,C)$. Particularly, Stable Diffusion (SD) represents a text-conditioned LD model, which uses text embeddings $C$, produced by a CLIP encoder \cite{radford2021learning}, to enable stochastic text-to-image (T2I) generation. Trained on the LAION-5B database \cite{schuhmann2022laion}, SD stands as a foundation open-source T2I model which is commonly utilized by the research community as a powerful image prior for downstream tasks.

\subsection{Base Models}
To demonstrate the effectiveness of ID-embeddings in face reconstruction and highlight the necessity of extensive datasets, we conducted initial experiments training an ID-conditioned model from scratch on low-resolution images with varying data sizes. This initial model follows the conditional LD format \cite{rombach2022high}, \ie a perceptual autoencoder and a UNet denoiser equipped with cross-attention layers, and is trained on image-embedding pairs using the standard loss of Eq.~\ref{eq:unet_loss}.

\begin{figure}
    \vspace{-0.3cm}
    \centering
    \begin{subfigure}{0.45\textwidth}
        \includegraphics[width=\linewidth, trim={2cm 0.3cm 2cm 0.3cm}, clip]{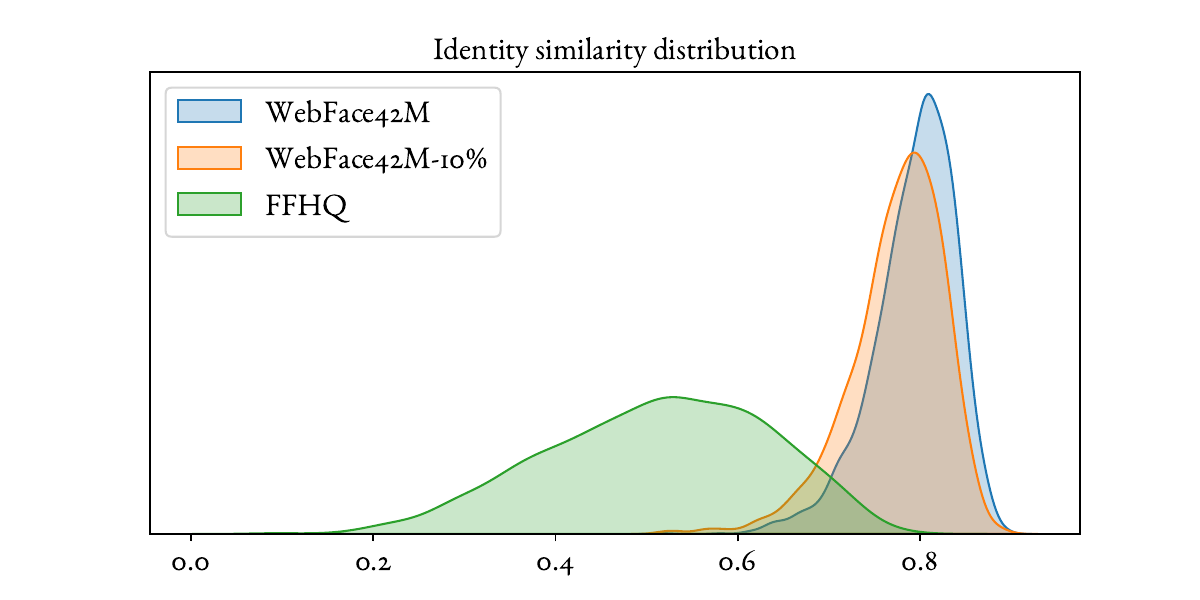}
        \caption{Synth-500}
    \end{subfigure}
    \hfill
    \begin{subfigure}{0.45\textwidth}
        \includegraphics[width=\linewidth, trim={2cm 0.3cm 1.8cm 0.3cm}, clip]{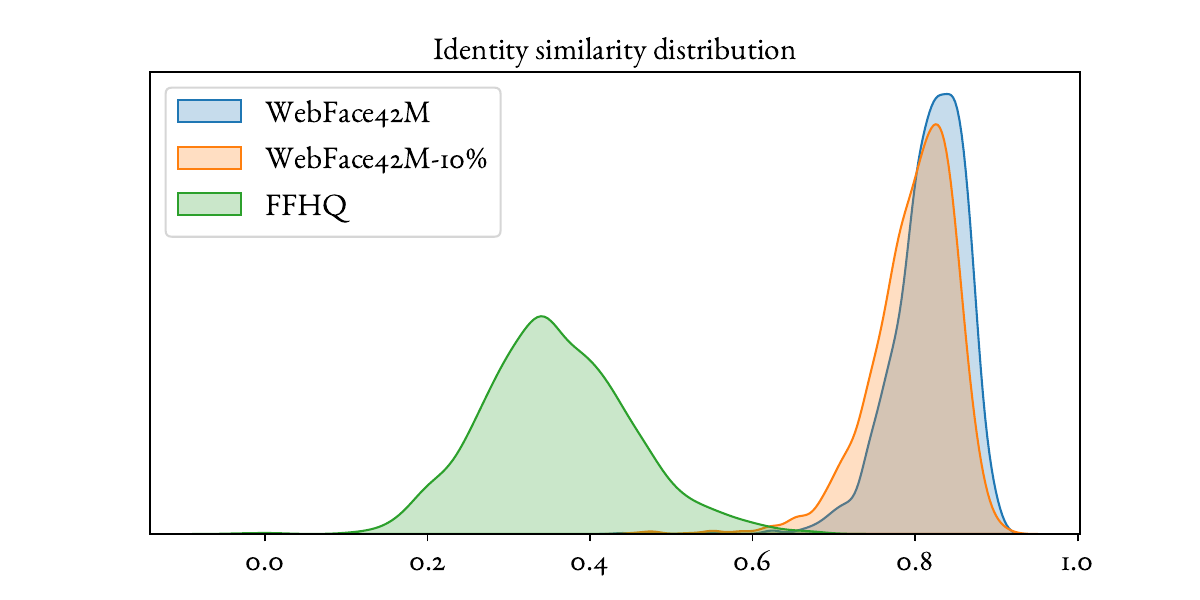}
        \caption{AgeDB}
    \end{subfigure}

    \caption{ArcFace~\cite{deng2019arcface} similarity distributions between input and generated faces from LD models trained on the ID-to-image task. We use two different datasets of input IDs for evaluation (500 and 400 IDs respectively) and generate 5 images per ID. We compare models trained on three datasets: FFHQ, WebFace42M-10\%, and WebFace42M.}
    \label{fig:id_sim_ldm}
    \vspace{-0.5cm}
\end{figure}

Fig.~\ref{fig:id_sim_ldm} compares the performance of the model when trained on three different datasets: FFHQ \cite{karras2019style} (70K single-person images, downsampled to $256 \times 256$), WebFace42M-10\% (a subset with $\sim$4M images from $\sim$200K identities), and the complete WebFace42M \cite{zhu2021webface260m} dataset. For each model, we generate images based on 400 real and 500 synthetic input faces (see Sec.~\ref{sec:datasets} for the images used in our experiments) and measure the ID similarity between the input and generations. Results reveal that a model trained on FFHQ \cite{karras2019style} exhibits limited ID retention due to the relatively modest number of images and IDs. In contrast, WebFace42M~\cite{zhu2021webface260m}, despite comprising low-resolution face images ($112\times 112$ pixels) cropped around the facial region for FR training, proves particularly suitable for our task, even when reduced to 10\% size. Notably, it constitutes the largest public FR dataset, consisting of approximately 42M images,
and, in contrast to other facial datasets (\eg~\cite{karras2019style, schuhmann2021laion, zheng2022general}), it contains significant and consistent intra-class variation, which is crucial for diverse generations of the same ID.

Expanding upon this finding, we then develop \modelname{}, which enables the synthesis of photo-realistic images at higher resolutions. This model is derived by fine-tuning SD on a restored version of WebFace42M \cite{zhu2021webface260m}, along with FFHQ \cite{karras2019style} and CelebA-HQ \cite{karras2017progressive} for enhanced quality, as detailed below. 

\subsection{\modelname{}}
Motivated by the ID retention offered by large-scale databases, we structure our model in two ways:
1) we employ a pre-trained SD~\cite{rombach2022high} as our prior, 2) we automatically generate a high-resolution dataset from~\cite{zhu2021webface260m}, enabling efficient training of our backbone without compromising its superior quality priors.

\subsubsection{ID-Conditioning:}

Our model is implemented with \texttt{stable-diffusion-v1-5}, which uses a CLIP text encoder~\cite{radford2021learning} to guide image synthesis. Our goal is to condition it on ArcFace embeddings, while directly harnessing the generative power of its UNet. Thus, it is necessary to project ArcFace embeddings to the space of CLIP embeddings, used by the original model. We achieve this by feeding them into the same encoder and fine-tuning it to swiftly adapt itself to the ArcFace input. This approach offers a more seamless projection than replacing CLIP with an MLP as in recent works~\cite{li2023photomaker, yan2023facestudio, wang2024stableidentity} (see Sec.~\ref{sec:mlp} for an ablation comparison). To ensure compatibility with CLIP, we employ a simple prompt, ``photo of a <id> person''. Following tokenization, we substitute the placeholder <id> token embedding with the ArcFace vector $\textbf{w}$, yielding a sequence of token embeddings $s = \{\textbf{e}_1, \textbf{e}_2, \textbf{e}_3, \hat{\textbf{w}}, \textbf{e}_5\}$. Here, $\hat{\textbf{w}}\in \mathbb{R}^{768}$ corresponds to $\textbf{w}\in \mathbb{R}^{512}$ after zero-padding to match the dimension of $\textbf{e}_i\in \mathbb{R}^{768}$. The resulting sequence is fed to the encoder $\tau$, which maps it to the CLIP output space $C = \tau(s) \in \mathbb{R}^{N\times 768}$ (with $N$ denoting the tokenizer's maximum sentence length). This process is illustrated in Fig.~\ref{fig:method}. Note that as a byproduct of this operation, the ID information is shared across multiple embeddings in the output of $\tau$, offering more detailed guidance to the UNet. This concept is also
used in recent works~\cite{valevski2023face0, chen2023dreamidentity}. During training, we consistently employ this default pseudo-prompt for all images. This intentional choice directs the encoder's attention solely to the ID vector, disregarding any irrelevant contextual information. Consequently, through extensive fine-tuning, we effectively transform the text encoder into a face encoder specifically tailored for projecting ArcFace embeddings into the CLIP latent space.

\subsubsection{Dataset:}
We retain WebFace42M~\cite{zhu2021webface260m} for its vast size and intra-class variability,
which, however, suffers from low-resolution data and a tight facial crop.
Even more, the pre-trained SD backbone is designed for a resolution of $512\times 512$ and a similar resolution is required to fine-tune \modelname{}.
To alleviate this, we meticulously upsample them using GFPGAN (v1.4), a state-of-the-art blind face restoration network \cite{wang2021gfpgan}, denoted as $\gamma$.
We perform degradation removal and $4\times$ upscaling to $448\times448$, so that $\hat{x} = \gamma(x)$. GFPGAN is a GAN-based upsampling method, with a strong face prior using both adversarial and ID-preserving losses, achieving crisp and faithful restoration. We follow this process on a large portion of the original database, given our computational limits, acquiring approximately 21M images of 1M identities at $448 \times 448$ pixels.
Using the restored images $\hat{x}$, we train \modelname{}. Albeit of higher quality, this dataset is still limited to a tightly cropped facial area for FR training. While it allows to learn a robust ID prior, a complete face image is usually preferred. Thus, we further fine-tune it on FFHQ \cite{karras2019style} and CelebA-HQ \cite{karras2017progressive}, which consist of less constrained face images. Our final model generates FFHQ-aligned images at $512\times 512$ pixels.

\section{Experiments}
\label{sec:exp}

\subsection{ID-Consistent Generation}
We perform extensive quantitative comparisons to evaluate the performance of recent ID-conditioned models in generating both diverse and faithful images of a subject. Details about the methods and comparison process are provided below.

\subsubsection{Methods} We compare against recent zero-shot methods that condition synthesis on identity information. Typically, these methods employ either CLIP image features or FR features to achieve tuning-free customization of SD on a given face.
In particular, we compare against the following open-source methods: \\
1) \textit{FastComposer} \cite{xiao2023fastcomposer}, which combines text with features extracted from the CLIP image encoder. It further uses a localized attention mechanism and proposes delayed subject conditioning during sampling for improved text editability. It is trained on FFHQ-wild~\cite{karras2019style}, automatically annotated with text prompts.\\
2) \textit{PhotoMaker} \cite{li2023photomaker}, which extracts CLIP features from one or a few images of a subject and combines them with text to condition the model. It is trained on a custom dataset with 112K images from 13K celebrities collected from the web.\\
3) \textit{IP-Adapter-FaceID (IPA-FaceID)} \cite{ye2023ip,ye2023ip-github}, which uses a decoupled attention mechanism for ID features in addition to text. Subsequent versions (\textit{IPA-FaceID-Plus} and \textit{Plusv2}) also use a combination of ID with CLIP image embeddings. \\
4) \textit{InstantID} \cite{wang2024instantid}, which extends \textit{IPA-FaceID} with an IdentityNet, akin to ControlNet, conditioned on FR embeddings and sparse landmarks. It is trained on LAION-Face~\cite{zheng2022general}, plus 10M automatically annotated images from the web.

\vspace{-0.3cm}
\subsubsection{Prompt Selection}
Since the aforementioned methods are tailored for text-driven synthesis, their ability to generate consistent IDs depends heavily on the input prompt. Although this allows creative stylizations, it demands prompt engineering and user inspection. To eliminate subjective bias, we suggest an automatic evaluation using a simple prompt, ``photo of a person'', for all samples. This enables us to assess ID-retention performance based on face features without elaborate text descriptions, ensuring a fair comparison across all methods.

\begin{figure}[t]
  \begin{minipage}[t]{0.75\textwidth}
    \centering
    \begin{subfigure}{0.47\linewidth}
      \includegraphics[width=\linewidth, trim={3.0cm 0.3cm 2.5cm 0.5cm}, clip]{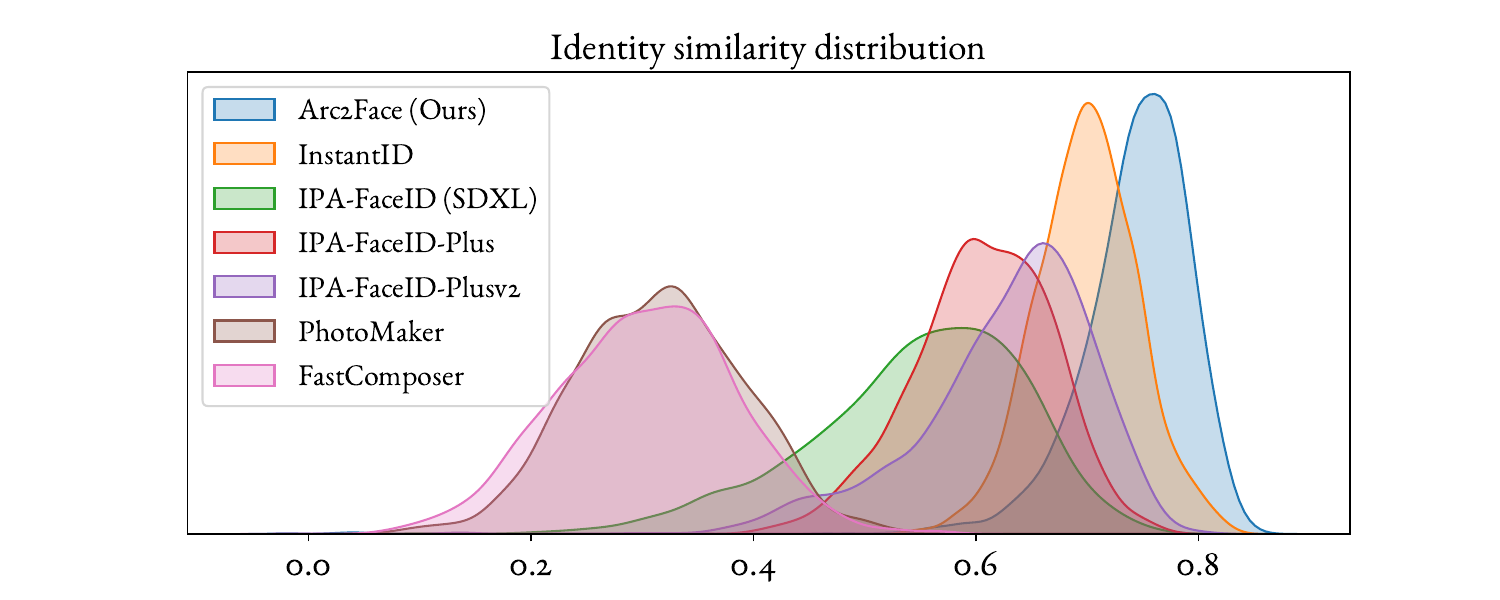}
        \caption{Synth-500}
    \end{subfigure}
    \begin{subfigure}{0.47\linewidth}
      \includegraphics[width=\linewidth, trim={3.0cm 0.3cm 2.3cm 0.5cm}, clip]{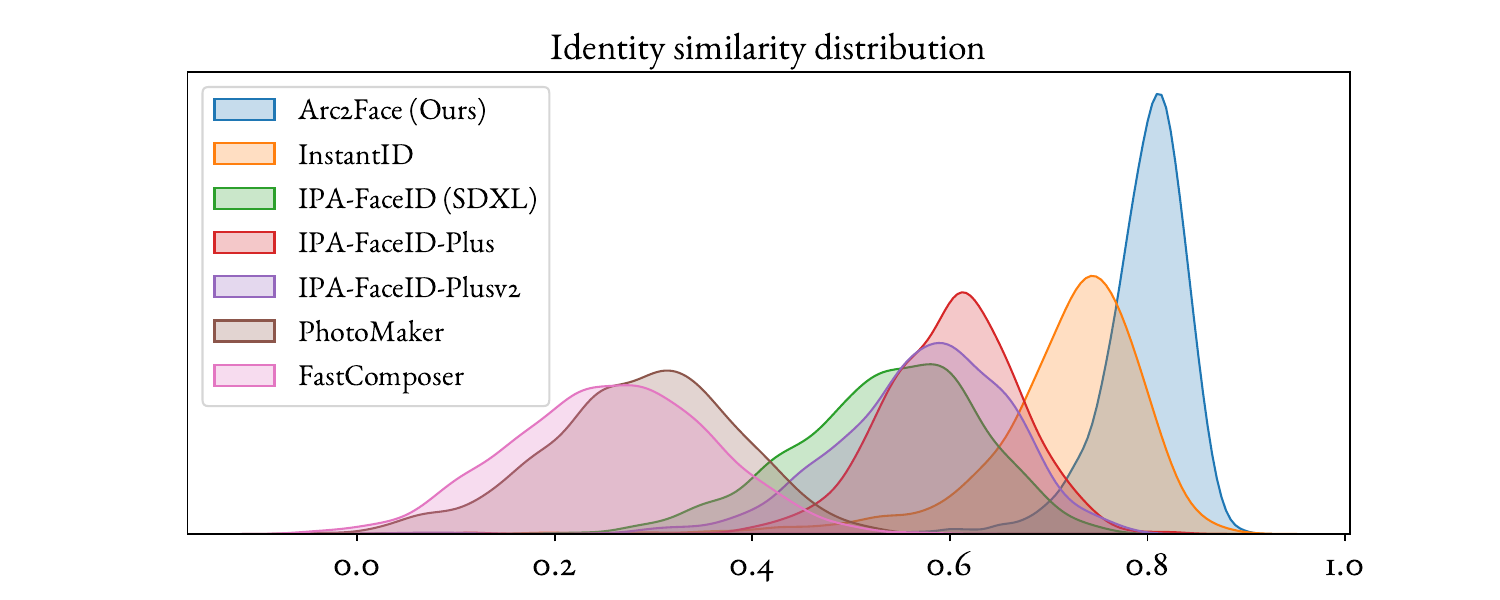}
        \caption{AgeDB}
    \end{subfigure}
    \caption{Distribution of ArcFace similarity between input IDs, synthetic (a) or real (b), and generated images of them by different models. As all non-CLIP-based methods use \cite{deng2019arcface} for conditioning, we evaluate them with~\cite{deng2019arcface}. For an evaluation with a different network, please refer to the Supp.~Material, where similar observations can be made.}
    \label{fig:id_sim}
  \end{minipage}%
  \hfill 
  \begin{minipage}[t]{0.2\textwidth}
    \centering
    \includegraphics[width=0.92\linewidth, trim={1.5cm -0cm 1.3cm 1cm}, clip]{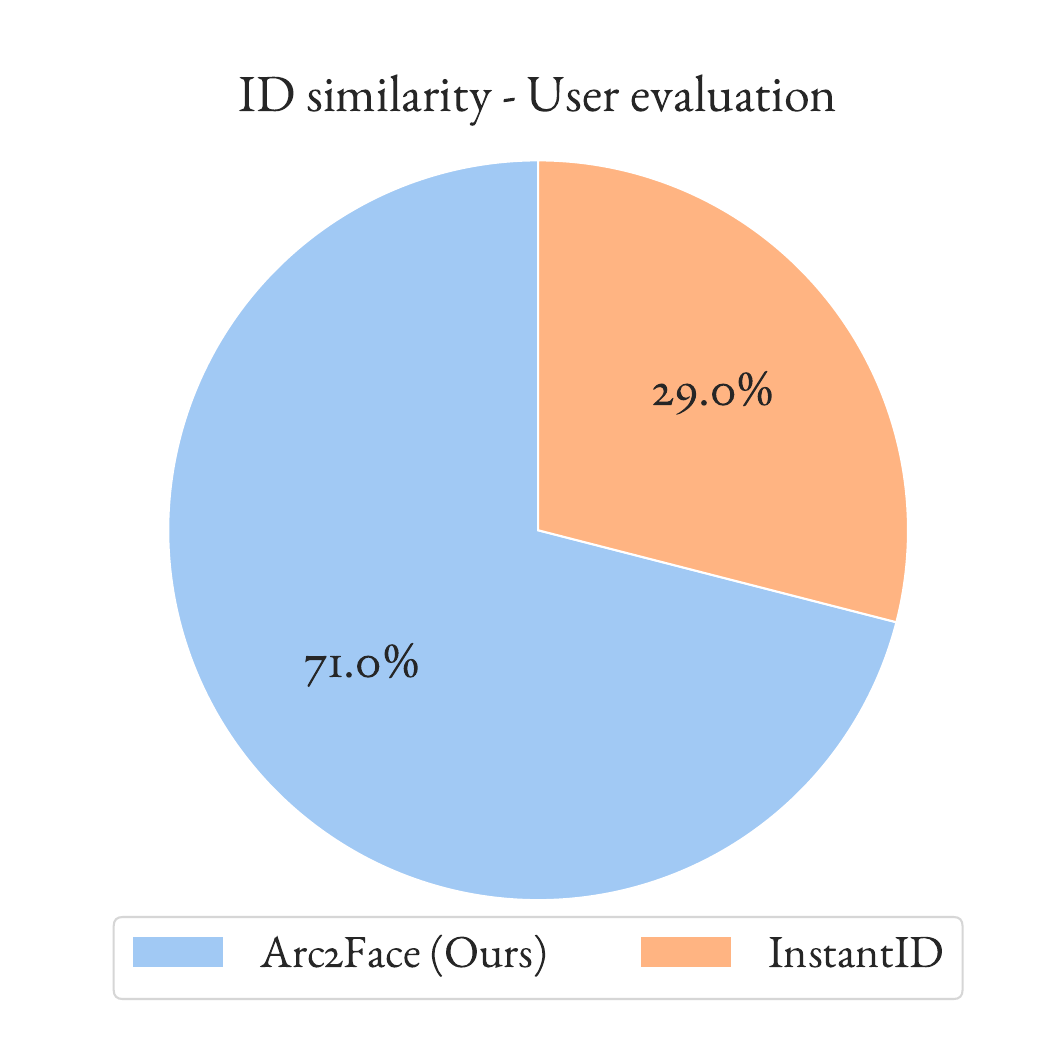}
      \caption{Percentage of user votes received by our method and InstantID re. ID fidelity.}
      \label{fig:user_study}
  \end{minipage}
  \vspace{-0.3cm}
\end{figure}

\vspace{-0.3cm}
\subsubsection{Datasets and Metrics} 
\label{sec:datasets}
We construct two datasets for our comparison. The first, hereafter denoted as Synth-500, includes 500 images of never-before-seen identities, generated by~\cite{tpdne}. The second is a selection of 400 real images from the public AgeDB~\cite{moschoglou2017agedb} database, chosen based on higher resolution. For each image, we generate 5 samples using all methods and calculate the following metrics: 1) ID similarity: we compute the cosine similarity of ID features between the input and generated faces. The corresponding distributions are plotted in Fig.~\ref{fig:id_sim}. 2) LPIPS distance: similarly to~\cite{li2023photomaker}, we assess the diversity of generation by calculating the pairwise perceptual distance (LPIPS) \cite{zhang2018unreasonable} between images of the same ID. The average distance for all pairs is reported in Tab~\ref{tab:benchmark}. To focus on the actual facial diversity, we first detect and align all faces and remove the background using an off-the-shelf parser~\cite{face-parsing-pytorch}. 3) Exp./Pose diversity: we predict the expression and pose (jaw/neck articulation) parameters of the FLAME model~\cite{FLAME:SiggraphAsia2017} for each sample using EMOCA v2~\cite{danvevcek2022emoca, filntisis2022visual} and compute the average pairwise $\ell_2$ expression and pose distances between images of the same ID. 4) FID: we use the FID~\cite{heusel2017gans, Seitzer2020FID} metric to assess the quality of samples (after face detection and alignment) with respect to the input images. Results are presented in Tab~\ref{tab:benchmark}.

\begin{table}
    \begin{center}
    \setlength{\tabcolsep}{2.1pt}
    \scriptsize
    \begin{tabular}{lcccccccc} 
        \toprule
         & \multicolumn{2}{c}{LPIPS$\uparrow$} & \multicolumn{2}{c}{Exp. ($\ell_2$)$\uparrow$} & \multicolumn{2}{c}{Pose ($\ell_2$)$\uparrow$} & \multicolumn{2}{c}{FID$\downarrow$}\\
         & Synth-500 & AgeDB & Synth-500 & AgeDB & Synth-500 & AgeDB & Synth-500 & AgeDB \\
         \midrule
         FastComposer & 0.389 & 0.487 & 3.597 & 4.678 & 0.163 & 0.225 & 13.517 & 31.736\\
         Photomaker & 0.410 & 0.424 & 3.920 & 4.283 & 0.167 & 0.165 & 13.295 & 8.410\\ 
         InstantID & 0.386 & 0.437 & 3.733 & 4.569 & 0.059 & 0.082 & 22.859 & 18.598\\ 
         IPA-FaceID (SDXL) & 0.402 & 0.462 & 4.648 & 5.812 & 0.181 & 0.197 & 7.104 & 24.105\\ 
         IPA-FaceID-Plus & 0.320 & 0.384 & 2.706 & 3.518 & 0.150 & 0.194 & 14.880 & 11.817\\ 
         IPA-FaceID-Plusv2 & 0.356 & 0.429 & 3.147 & 4.092 & 0.185 & 0.236 & 9.752 & 10.798\\
         \textbf{\modelname{} (Ours)} & \textbf{0.506} & \textbf{0.508} & \textbf{6.375} & \textbf{5.966} & \textbf{0.317} & \textbf{0.273} & \textbf{5.673} & \textbf{6.628}\\
        \bottomrule
    \end{tabular}
    \end{center}
    \vspace{-0.2cm}
    \caption{Quantitative comparison between \modelname{} and~\cite{xiao2023fastcomposer, li2023photomaker, wang2024instantid, ye2023ip} on 500 synthetic and 400 real IDs. We produce 5 samples per ID for all methods and assess the diversity of generated faces using perceptual and 3DMM-based distances, as well as their quality based on FID. Bold values denote the best results in each metric.}
    \label{tab:benchmark}
    \vspace{-0.8cm}
\end{table}

\begin{figure*}[t]
    \centering
    \includegraphics[width=0.85\textwidth]{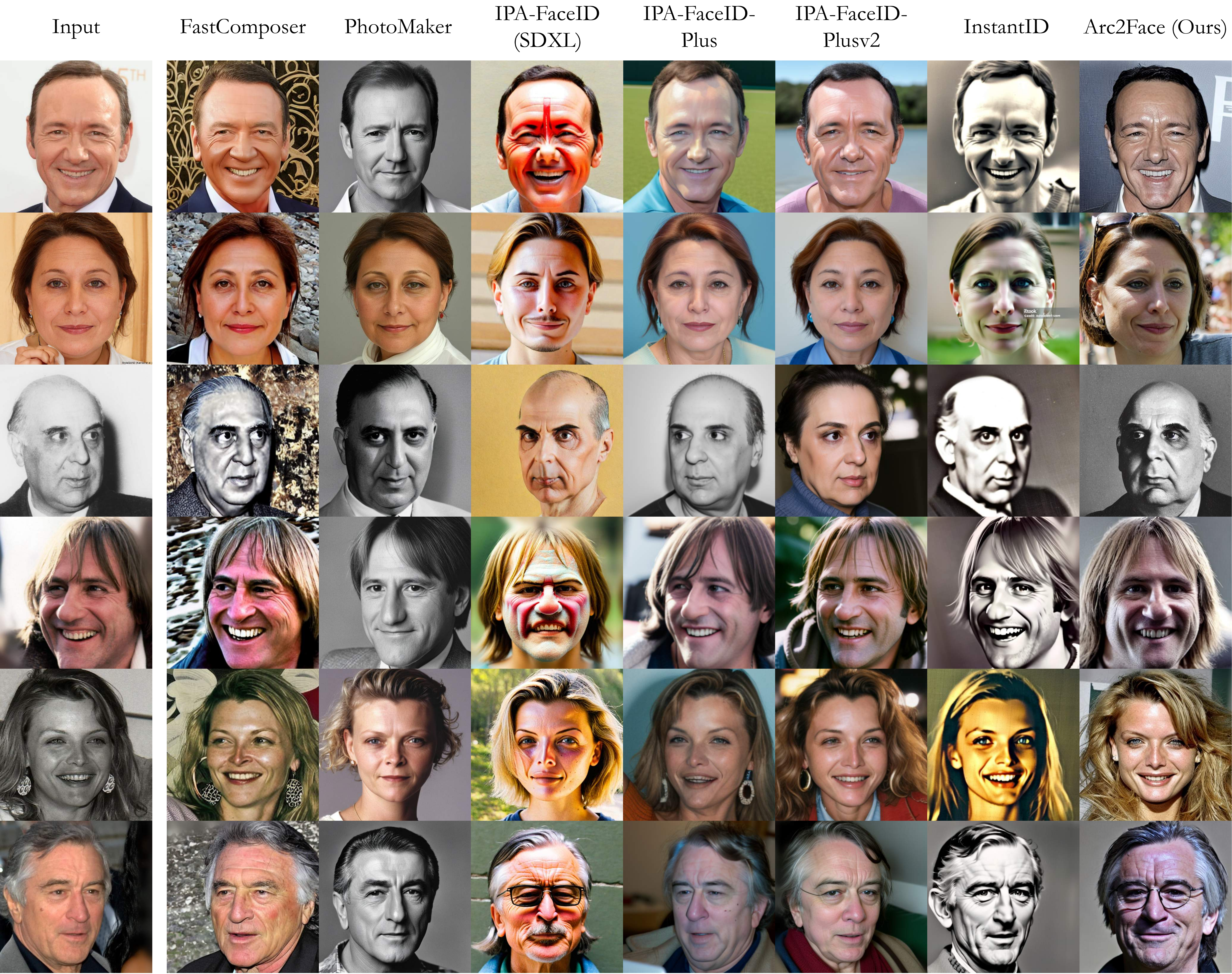}
    \caption{Visual comparison of \modelname{} with state-of-the-art methods~\cite{xiao2023fastcomposer, li2023photomaker, ye2023ip, wang2024instantid}  using the abstract prompt ``photo of a person'' to focus on their ID-conditioning ability.}
    \label{fig:comp}
\end{figure*}

As depicted in Fig.~\ref{fig:id_sim}, existing methods, particularly \cite{li2023photomaker, xiao2023fastcomposer} that rely on CLIP 
features, struggle to preserve the ID without detailed text descriptions of the subject. InstantID~\cite{wang2024instantid} attains the highest similarity, behind \modelname{}. However, this results in limited pose/exp. diversity as it further uses the input facial landmarks to constrain generation. \modelname{} does not require any text or spatial conditions and achieves the highest facial similarity as well as diversity and realism across both datasets. This particularly highlights the efficiency of ID-embeddings in the context of face generation against CLIP image or text features. A visual comparison with the above methods is provided in Fig.~\ref{fig:comp}, whereas additional qualitative results and visualizations are provided in the Supp.~Material. We further conducted a user study to compare with the second best-performing method in terms of ID similarity, \ie InstantID~\cite{wang2024instantid}. In particular, we asked 50 users to choose the method whose result best resembles the input face, regardless of quality or realism, for a randomly selected set of 30 IDs from our datasets. The two methods were randomly presented side-by-side. Fig.~\ref{fig:user_study} confirms a strong preference for \modelname{} in terms of ID resemblance.

\subsection{Face Recognition with Synthetic Data}

To further demonstrate the potential of our model, we use it to create synthetic images for FR training. In particular, we sample novel ID vectors from the distribution of ArcFace embeddings learned by PCA on WebFace42M~\cite{zhu2021webface260m}, while ensuring sufficient intra-class variance for the vectors of the same ID. To guarantee the uniqueness of generated subjects, we keep only those with ID similarity between them below $0.3$. We compare against recent methods that train FR models on synthetic data, namely SynFace~\cite{qiu2021synface}, DigiFace~\cite{bae2022digiface}, and DCFace~\cite{kim2023dcface}, following their training scheme with IR-SE-50~\cite{deng2019arcface} as a backbone and the AdaFace~\cite{kim2022adaface} loss function. The trained models are evaluated on LFW~\cite{lfw}, CFP-FP~\cite{cfpfp}, CPLFW~\cite{cplfw}, AgeDB~\cite{moschoglou2017agedb} and CALFW\cite{calfw} datasets, ensuring 
large pose~\cite{cfpfp, cplfw} and age~\cite{moschoglou2017agedb, calfw} variation. In Tab.~\ref{tab:FRcomparison}, we present results for $0.5$M and $1.2$M settings, corresponding to the size of CASIA-WebFace~\cite{yi2014learning} and an increased size, respectively. \modelname{} surpasses DCFace in all five test datasets with an average improvement of $2.17\%$ in the $0.5$M regime and $1.93\%$ in the $1.2$M regime.
Especially on the CFP-FP dataset, \modelname{} significantly outperforms DigiFace and DCFace, showing high ID-consistency under large pose variation.   

\begin{table*}
\vspace{-0.18cm}
\begin{center}
\scriptsize
\scalebox{0.9}{
\begin{tabular}{l|c|l|c|c|c|c|c||c}
\hline
    Methods & Venue  &\# images (\# IDs$\!\times\!$ \# imgs/ID) & LFW & CFP-FP & CPLFW & AgeDB & CALFW & Avg  \\ 
    \hline\hline
    SynFace & ICCV21  & $0.5$M ($10$K $\!\times\!$ $50$)  & $91.93$    & $75.03$    & $70.43$ & $61.63$ & $74.73$ & $74.75$   \\ 
    DigiFace & WACV23 & $0.5$M ($10$K $\!\times\!$ $50$)  & $95.4$     & $87.4$     & $78.87$ & $76.97$ & $78.62$ & $83.45$   \\ 
    DCFace   & CVPR23 & $0.5$M ($10$K $\!\times\!$ $50$)  &  $98.55$   &  $ 85.33 $ &  $82.62$ &  $89.70$ &  $91.60$ &  $89.56$  \\
    \modelname{} & -      & $0.5$M ($10$K $\!\times\!$ $50$)  &  $\bm{98.81}$  & $\bm{91.87}$  & $\bm{85.16}$  &  $\bm{90.18}$ & $\bm{92.63}$ & $\bm{91.73}$  \\
    \hline 
    DigiFace & WACV23  & $1.2$M ($10$K $\!\times\!$ $72$ + $100$K $\!\times\!$ $5$) & $96.17$  & ${89.81}$ &  $82.23$ &  $81.10$ &  $82.55$  & $86.37$   \\ 
    DCFace   &CVPR23   & $1.2$M ($20$K $\!\times\!$ $50$ + $40$K $\!\times\!$ $5$)  &  $98.58$  & $88.61$ &  $85.07$ &  $90.97$  & $92.82$  & $91.21$   \\ 
    \modelname{} & -       & $1.2$M ($20$K $\!\times\!$ $50$ + $40$K $\!\times\!$ $5$)  &  $\bm{98.92}$  & $\bm{94.58}$  & $\bm{86.45}$  & $\bm{92.45}$  & $\bm{93.33}$ & $\bm{93.14}$  \\
    \hline
    \multicolumn{2}{c|}{CASIA-WebFace (Real)}  & $0.49$M (approx. $10.5$K $\!\times\!$ $47$)  &  $99.42$ &  $96.56$ &  $89.73$  & $94.08$  & $93.32$ &  $94.62$   \\ \hline
\end{tabular}
}
\end{center}
\vspace{-0.19cm}
\caption{Verification accuracies of FR models trained with synthetic datasets. SynFace~\cite{qiu2021synface} is a GAN-based dataset with a latent space mixup technique. DigiFace~\cite{bae2022digiface} is a 3D model-based dataset with image augmentation. DCFace~\cite{kim2023dcface} is a diffusion-based dataset with separate ID and style as dual conditions.
}
\label{tab:FRcomparison}
\vspace{-0.9cm}
\end{table*}

\vspace{-0.4cm}
\subsection{Pose/Expression Control}

\begin{figure}
    \centering
    \includegraphics[width=0.76\textwidth]{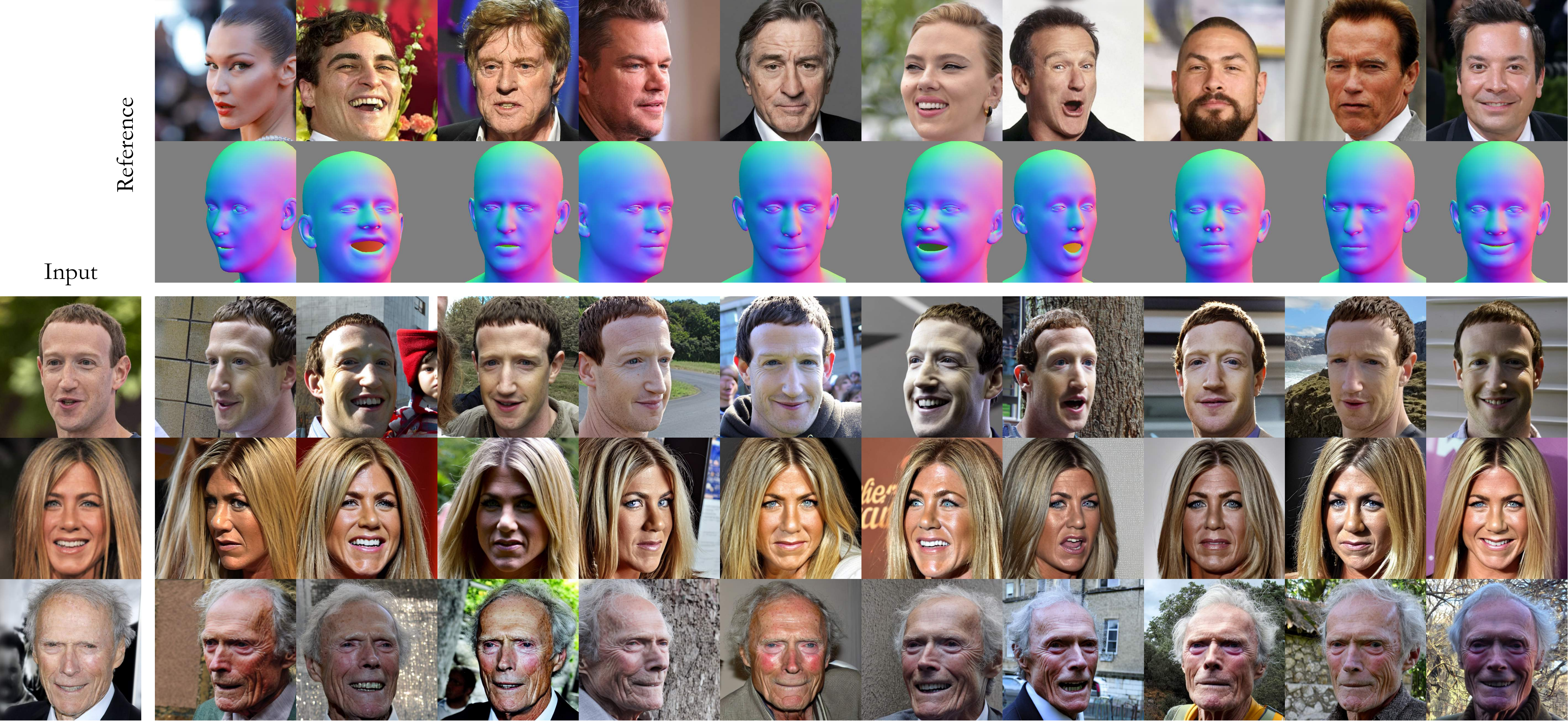}
    \caption{Samples from \modelname{}, conditioned on a 3DMM~\cite{FLAME:SiggraphAsia2017, danvevcek2022emoca} using ControlNet~\cite{zhang2023adding}.}
    \label{fig:controlnet}
    \vspace{-0.2cm}
\end{figure}

Our model can be trivially combined with ControlNet~\cite{zhang2023adding} for spatial control of the output. In particular, we use EMOCA v2~\cite{danvevcek2022emoca, filntisis2022visual} to perform 3D reconstruction on FFHQ \cite{karras2019style}, and train a ControlNet module, conditioned on the rendered face normals. During inference, we can render the 3D face normals of a source person under the expression and pose extracted from a reference image. This rendering is used to guide the synthesis of the source identity as shown in Fig.~\ref{fig:controlnet}. 

\subsection{Ablation Studies}
\label{sec:exp_ablation}

\begin{figure}
\vspace{-0.2cm}
  \begin{minipage}[t]{0.5\textwidth}
    \centering
    \begin{subfigure}{0.47\linewidth}
      \includegraphics[width=\linewidth, trim={0.8cm 0.5cm 0.8cm 0.5cm}, clip]{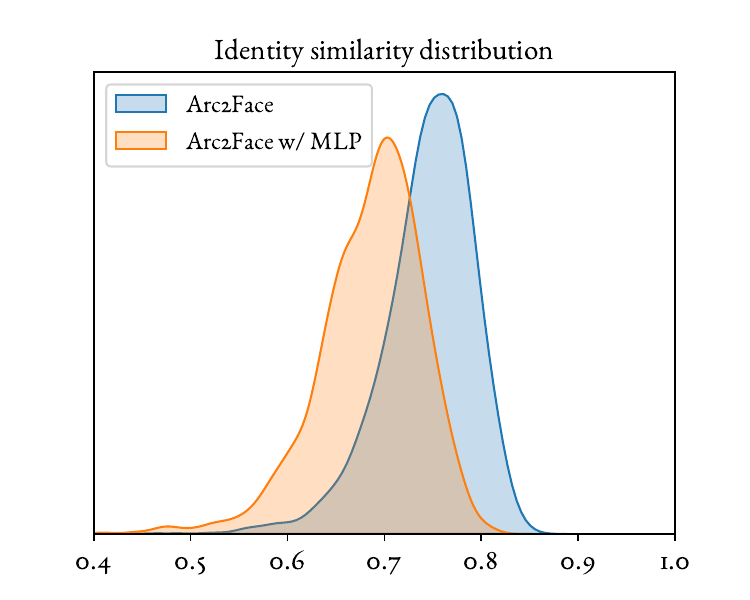}
      \caption{Synth-500}
    \end{subfigure}
    \begin{subfigure}{0.47\linewidth}
      \includegraphics[width=\linewidth, trim={0.8cm 0.5cm 0.8cm 0.5cm}, clip]{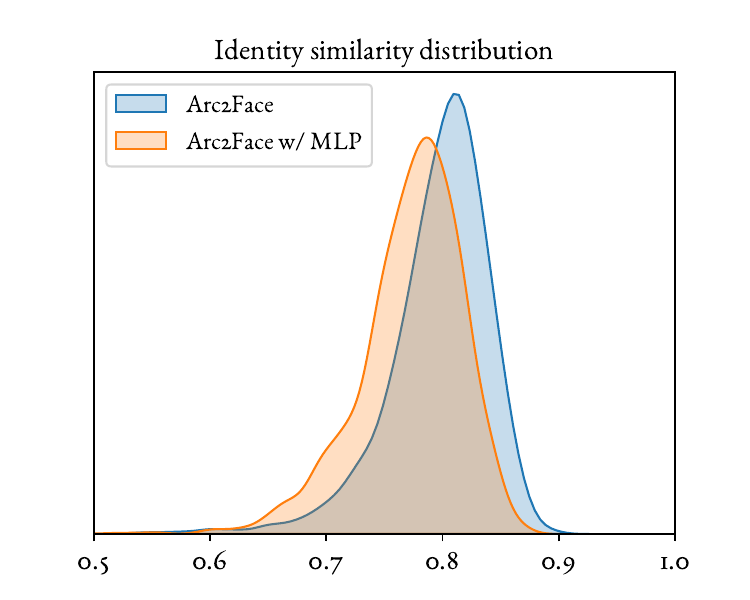}
      \caption{AgeDB}
    \end{subfigure}
    \caption{ID similarity \cite{deng2019arcface} distributions between input and generated images from our model. We compare the proposed approach with that of using an MLP for the projection of ID vectors to the CLIP latent space.}
    \label{fig:id_sim_ablation}
  \end{minipage}%
  \hfill
  \begin{minipage}[t]{0.48\textwidth}
    \centering
    \begin{subfigure}[b]{0.47\linewidth}
      \includegraphics[width=\linewidth, trim={0cm 0cm 0.8cm 0.5cm}, clip]{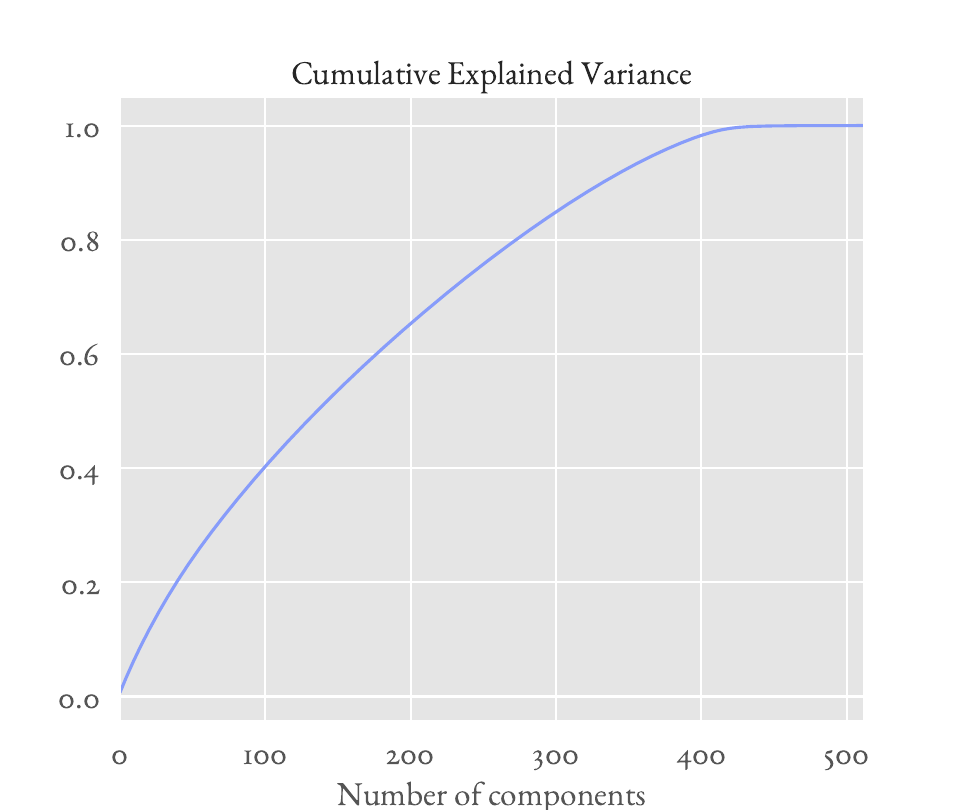}
      \caption{Cumulative Var.}
      \label{fig:pca_curve}
    \end{subfigure}
    \begin{subfigure}[b]{0.47\linewidth}
      \includegraphics[width=\linewidth, trim={0cm -8cm 0cm 0cm}, clip]{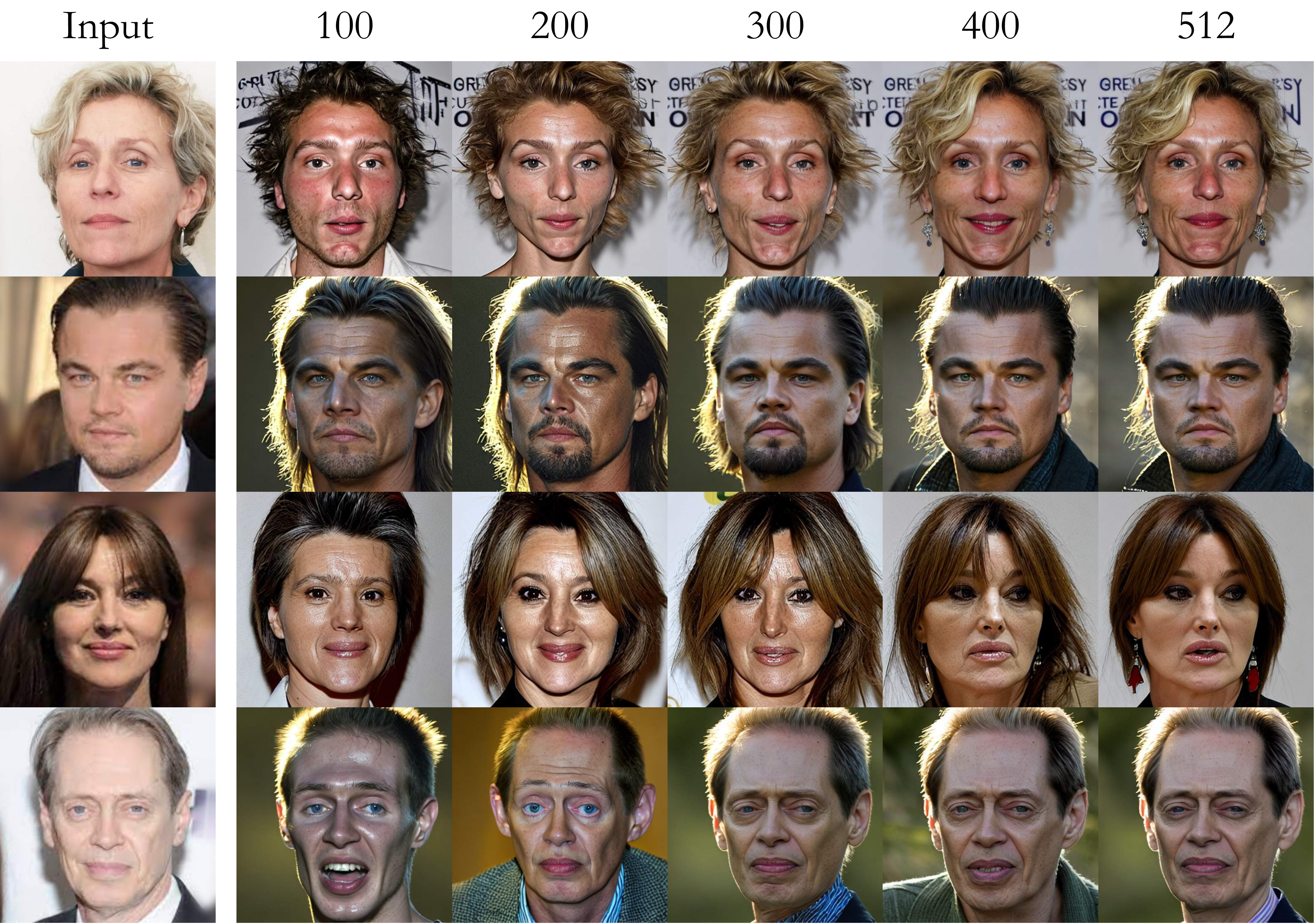}
      \caption{PCA projections}
      \label{fig:pca_n_comp}
    \end{subfigure}
    \caption{PCA on ID-embeddings: We show the cumulative percentage of variance across components (a) and samples from \modelname{} when projecting ID vectors to a varying number of components (b).}
    \label{fig:pca}
  \end{minipage}
\end{figure}

\noindent\textbf{ID-Conditioning via MLP}\quad
\label{sec:mlp}
To achieve ID-conditioning in SD, we map ArcFace~\cite{deng2019arcface} features to the CLIP embedding space by adapting the pre-trained text encoder~\cite{radford2021learning} through extensive fine-tuning. Recent approaches combining text and ID-conditioning typically employ a simple MLP for subject embedding projection~\cite{li2023photomaker, valevski2023face0, chen2023dreamidentity, yan2023facestudio, wang2024stableidentity, peng2023portraitbooth, xiao2023fastcomposer}. To validate our choice, we conducted an ablation study, replacing the CLIP encoder with a 4-layer MLP while maintaining the same training setting. Fig.~\ref{fig:id_sim_ablation} shows that our MLP-based model yields lower face similarity in generated images. Since SD is trained with CLIP embeddings, using the same encoder is a more natural choice than training an MLP to learn the CLIP latent space. Additionally, the former represents a well-established architecture, while the latter would necessitate a more extensive hyperparameter search, rather than a simple selection of a shallow MLP, as seen in most works.

\noindent\textbf{Principal Component Analysis}\quad
To explore the intrinsic dimensionality of our facial ID representation, we conducted PCA on the ArcFace embeddings derived from WebFace42M images. Fig. \ref{fig:pca_curve} depicts the cumulative percentage of variance explained by the principal components. Notably, maintaining facial fidelity requires at least 300-400 components, as fewer components result in noticeable distortion, as shown in Fig.~\ref{fig:pca_n_comp}. This reveals the challenge of compressing ArcFace embeddings significantly, emphasizing the inherent complexity of our facial ID representation.

\begin{figure}
    \vspace{-0.4cm}
    \centering
    \includegraphics[width=0.8\textwidth]{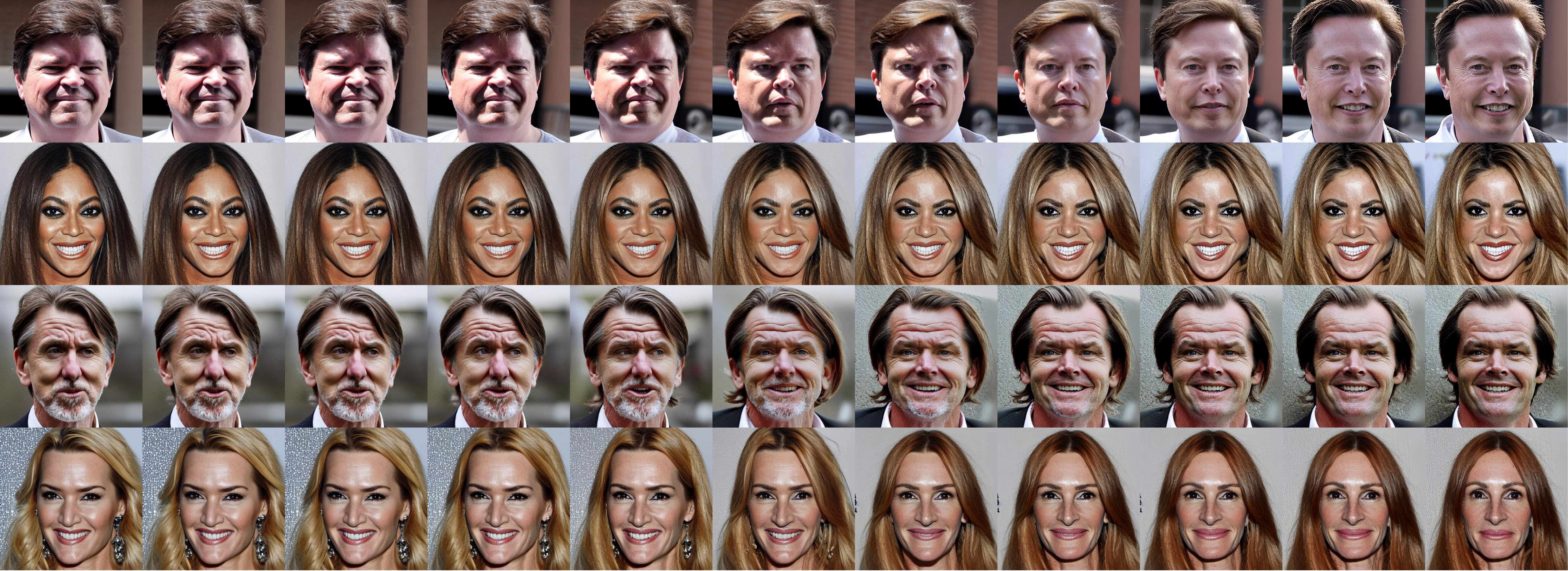}
    \caption{ID interpolations between pairs of subjects. \modelname{} generates plausible faces along the trajectory connecting their ArcFace vectors.}
    \label{fig:interpolations}
    \vspace{-0.4cm}
\end{figure}

\noindent\textbf{Averaging ID Features}\quad
ID-embeddings offer a compact representation of the facial characteristics that allows interpolation across different images. In Fig.~\ref{fig:interpolations}, we provide transitions between pairs of subjects by linearly blending their ArcFace vectors. In the Supp.~Material, we also show how ID resemblance may benefit from averaging embeddings from multiple images of the same person.

\begin{figure}
\vspace{-0.7cm}
    \centering
    \begin{subfigure}[b]{0.47\textwidth}
       \includegraphics[width=\linewidth, trim={2.8cm 0.3cm 2.5cm 0.3cm}, clip]{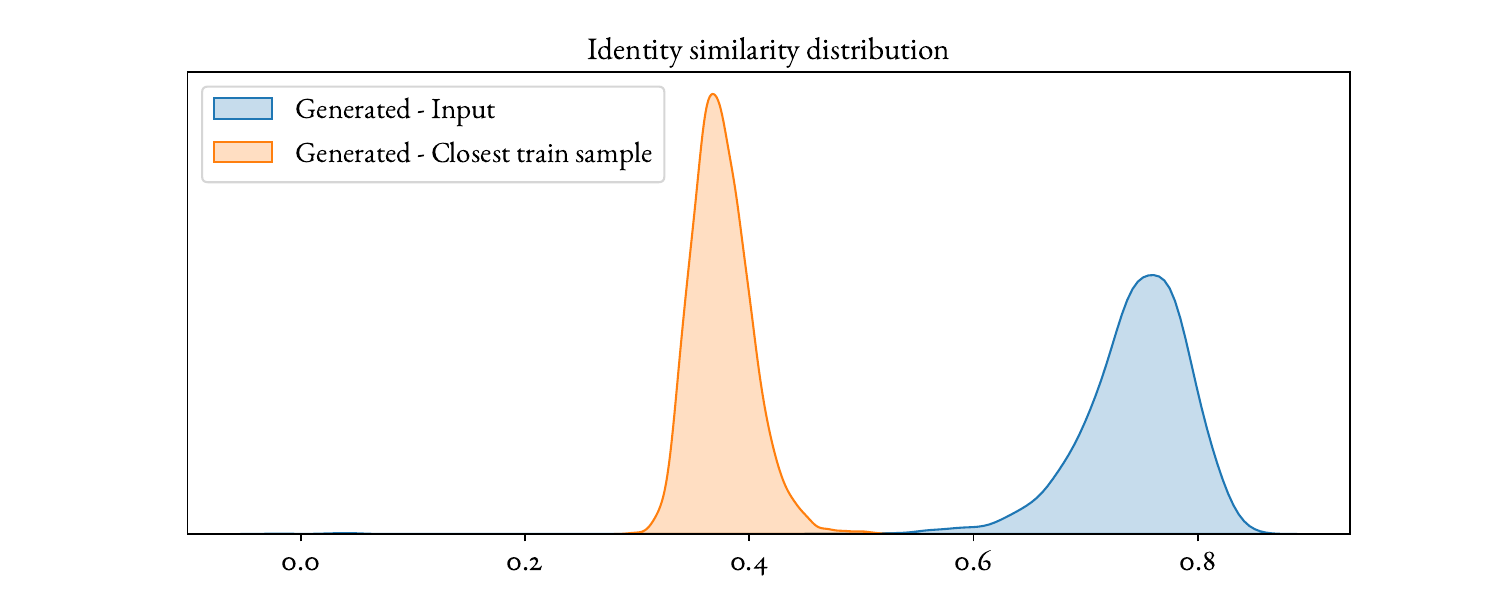}
        \caption{ArcFace similarity distributions between input/output and output/closest-train-image.}
        \label{fig:closest_train_sample_sim}
    \end{subfigure}
    \hfill
    \begin{subfigure}[b]{0.47\textwidth}
        \includegraphics[width=\linewidth, trim={0cm -2cm 0cm 0cm}, clip]{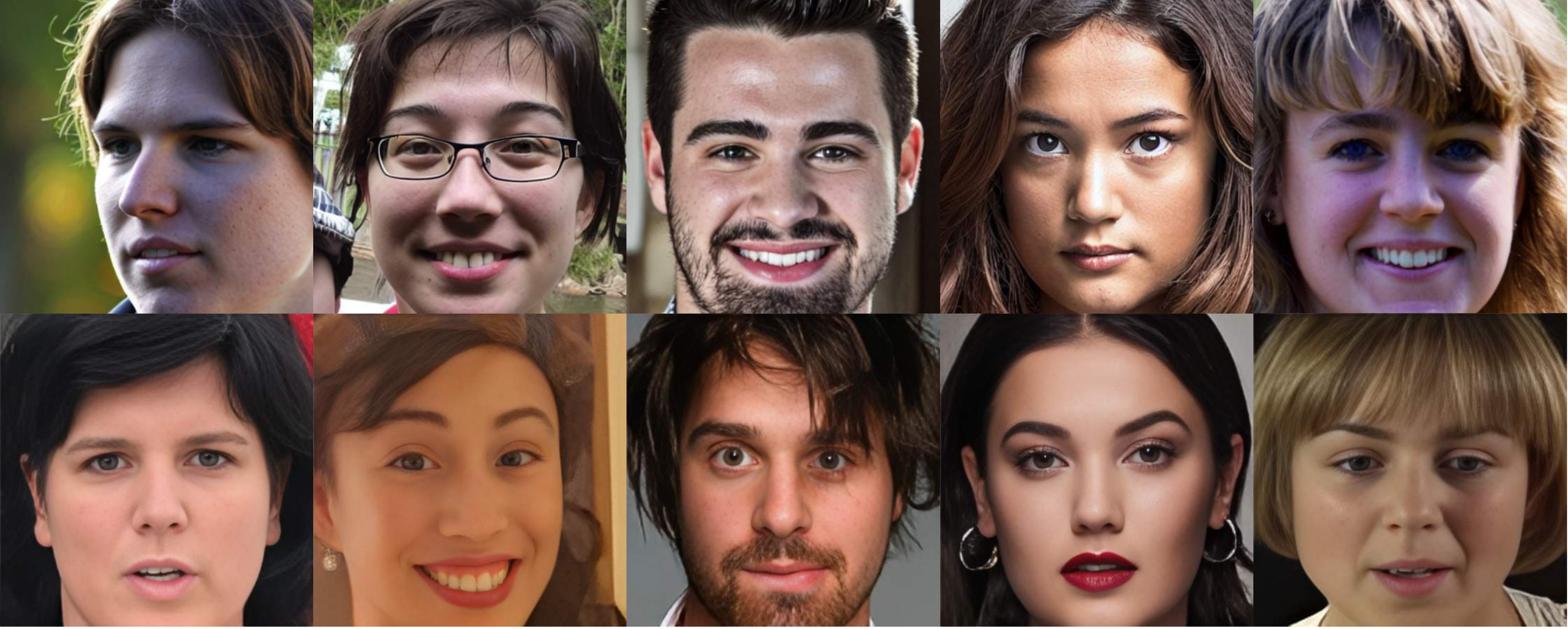}
        \caption{Examples of generated faces (top) and their closest train samples (bottom).}
        \label{fig:closest_train_samples}
    \end{subfigure}
    \caption{We show that \modelname{} does not replicate its training data. We generate images for 500 unseen IDs from Synth-500 and retrieve the train image with the highest ID similarity for each one. The average similarity between each output and its closest match is 0.37, whereas the similarity between the output and input features is 0.74.}
    \vspace{-0.6cm}
\end{figure}

\noindent\textbf{Generalization Ability}\quad
By training on an extensive dataset with more than 1M IDs and significant intra-class variance, our model is capable of reproducing photos of any person with high fidelity. To further assess its capacity, we examine whether or not \modelname{} has learned to replicate its training IDs by exploring the output images for the 500 previously unseen identities from Synth-500. In particular, for each of the 2.5K samples (5 per ID), we identified the closest training image - defined as the one exhibiting the highest ID similarity within the combined set of restored WebFace42M, FFHQ, and CelebA-HQ. Fig.~\ref{fig:closest_train_samples} shows some examples of generated images and their closest matches from the training data. We further plot the distribution of similarities in Fig.~\ref{fig:closest_train_sample_sim}. Notably, the generated images exhibit an average cosine similarity of 0.37 to their closest training sample and 0.74 to their (unseen) input features. This observation indicates that our model indeed does not memorize images from the training set, confirming its ability to faithfully reconstruct faces of novel identities.

\vspace{-0.3cm}
\subsection{Implementation Details}
\label{sec:exp_implementation}
We start from \texttt{stable-diffusion-v1-5} and fine-tune both the UNet and CLIP encoder with AdamW~\cite{loshchilov2017decoupled} and a learning rate of 1e-6, using 8 NVIDIA A100 GPUs and a batch size of 4 per GPU. First, we train on 21M restored images from WebFace42M for 5 epochs with a resolution of $448\times448$, and then fine-tune on $512\times512$ - sized FFHQ and CelebA-HQ images for another 15 epochs. All our results shown in this paper are generated using DPM-Solver~\cite{lu2022dpm, lu2022dpm++} with 25 inference steps and a classifier-free guidance scale of 3, which we empirically found to produce highly realistic images. For ID-embedding extraction, we use a frozen IR-100 ArcFace~\cite{deng2019arcface} trained on WebFace42M and normalize embeddings to unit magnitude. Regarding the methods we compare against, we use the official implementations and, in all cases, select the default hyper-parameters.

\vspace{-0.3cm}
\section{Limitations and Impact}
In this work, we introduce a model capable of generating high-quality facial images from facial embeddings.
Our method is limited by the capacity of the face embeddings encoder \cite{deng2019arcface}, as well as the datasets used \cite{zhu2021webface260m}, 
both of which are fortunately state-of-the-art.
Moreover, only one person per image can be generated.
Despite these limitations, we provide a foundation model, that can be further fine-tuned
to other datasets and modalities, aiding follow-up research.

\modelname{} can benefit various industrial and research applications,
including media, entertainment, and data generation for face analysis and synthesis.
We stress the ethical aspects of our work, as such technologies can be misused to create unoriginal facial images.
We thus restrict the training data to the facial region only.
The community must adhere to ethical responsibilities
and support fake content detection \cite{sun2023contrastive}.
Moreover, as such technologies proliferate, it is important to ensure that they remain equally effective
across different demographics.
Most prior art (\eg~\cite{li2023photomaker, wang2024stableidentity}), 
is based on biased celebrity datasets \cite{radford2021learning}. Instead, we use ID-embeddings, which although not perfect \cite{yucer2022measuring},
enable a general approach to facial generation
and encourage working with balanced synthetic datasets.

\vspace{-0.3cm}
\section{Conclusion}
\label{sec:conclusion}
In this work, we explore the power of ID features derived from FR networks as a comprehensive representation for face generation in the context of large-scale diffusion models. We 
show that million-scale face recognition datasets are required to effectively train an ID-conditioned model. To that end, we fine-tune the pre-trained SD on carefully restored images from WebFace42M. Our ID-conditioning mechanism transforms the model into an ArcFace-to-Image model, deliberately disregarding text information in the process. Our experiments demonstrate its ability to faithfully reproduce the facial ID of any individual, generating highly realistic images with a greater degree of similarity compared to any existing method, all while preserving diversity in the output. We hope our model encourages further research in ID-preserving generative AI for human faces.
\\
\textbf{Acknowledgements.} S. Zafeiriou and part of the research was funded by the EPSRC Fellowship DEFORM (EP/S010203/1) and EPSRC Project GNOMON (EP/X011364/1). B. Kainz received support from the ERC, project MIA-NORMAL 101083647, DFG 512819079, and by the State of Bavaria (HTA). HPC resources were provided by the Erlangen National High Performance Computing Center (NHR@FAU) of the Friedrich-Alexander-Universität Erlangen-Nürnberg (FAU) under the NHR project b180dc. NHR@FAU hardware is partially funded by the German Research Foundation (DFG) – 440719683.

%
%
\bibliographystyle{splncs04}
\bibliography{main}

\clearpage
\appendix

\begin{center}
\vspace*{8pt}
\textbf{\Large \modelname{}: A Foundation Model for ID-Consistent Human Faces}

\vspace{3pt}
\textbf{\Large (Supplementary Material)}
\vspace{20pt}
\end{center}

\section{Effect of Training Dataset Size}
\begin{figure}
    \centering
    \includegraphics[width=\textwidth]{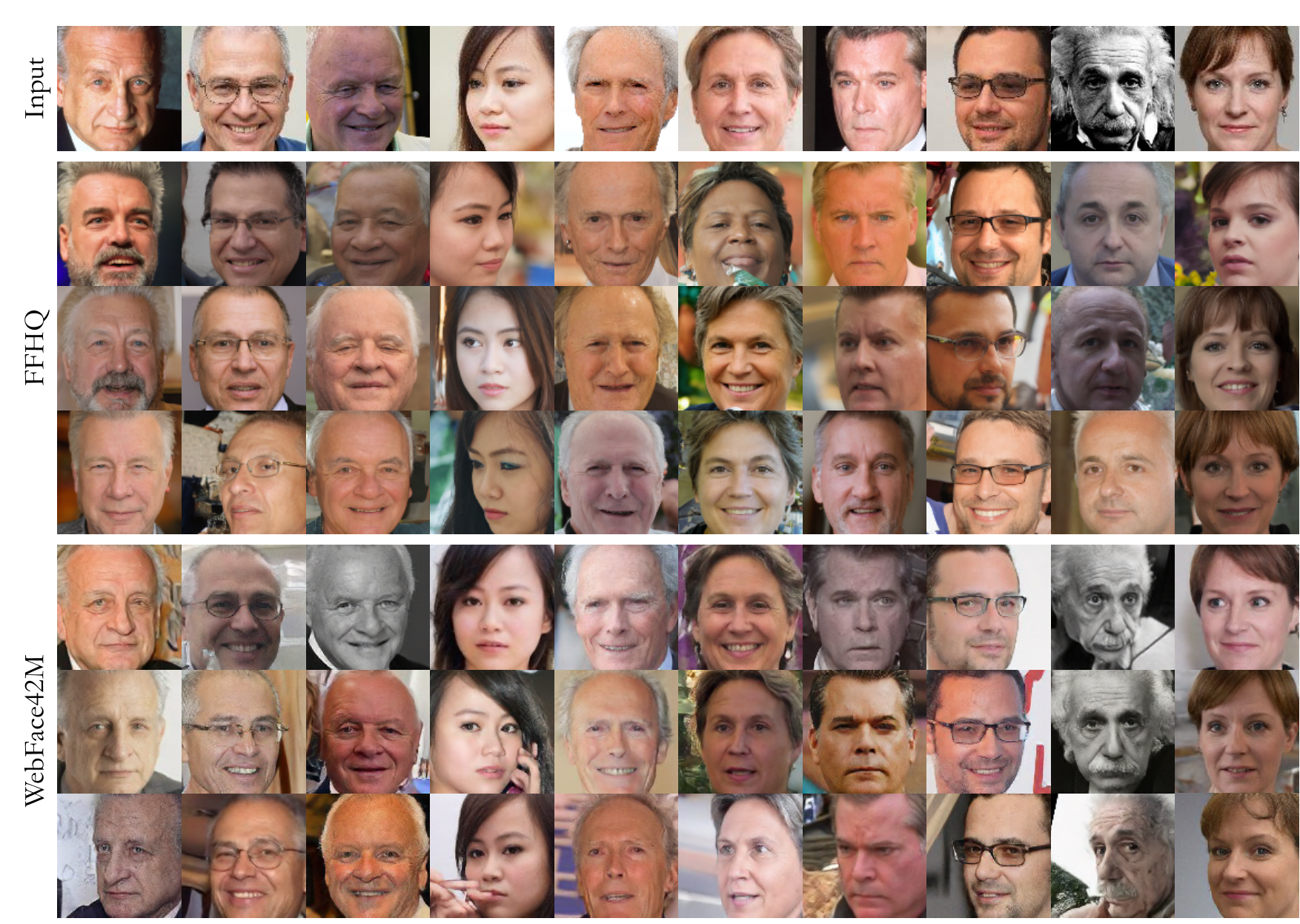}
    \caption{Visual comparison of ID-conditioned latent diffusion models trained from scratch on FFHQ and WebFace42M. FFHQ samples are aligned to the WebFace42M template ($112\times 112$ pixels) for simplicity.}
    \label{fig:low_res}
\end{figure}

As stated in the main paper, our initial experiments on low-resolution images showed that FFHQ~\cite{karras2019style} is insufficient for training a foundation ID-conditioned model, and large-scale datasets with multiple images per ID are essential for this task. In addition to the evaluation presented in Fig. \ref{fig:id_sim_ldm}, Fig.~\ref{fig:low_res} provides samples from our base Latent Diffusion model, trained on either FFHQ or WebFace42M~\cite{zhu2021webface260m} (before restoration), where significant ID-distortion can be observed for the former, confirming the superiority of FR datasets.

\section{Using Multiple Images per ID}

\begin{figure}
  \begin{minipage}[t]{0.62\textwidth}
    \centering
    \includegraphics[width=\linewidth]{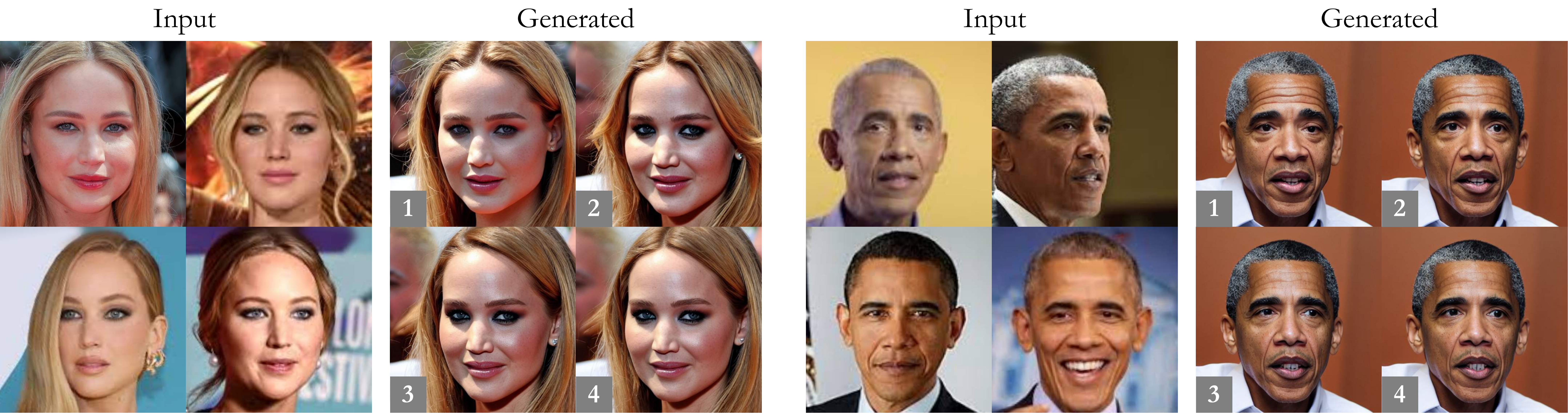}
      \caption{Effect of averaging ArcFace features from multiple input images. Increasing the number of ID vectors used (indicated at the bottom left of each result) leads to improved fidelity of samples.}
      \label{fig:multiple_images}
  \end{minipage}
  \hfill 
  \begin{minipage}[t]{0.33\textwidth}
    \centering
    \includegraphics[width=\linewidth, trim={3.0cm 0.3cm 2.5cm 0.5cm}, clip]{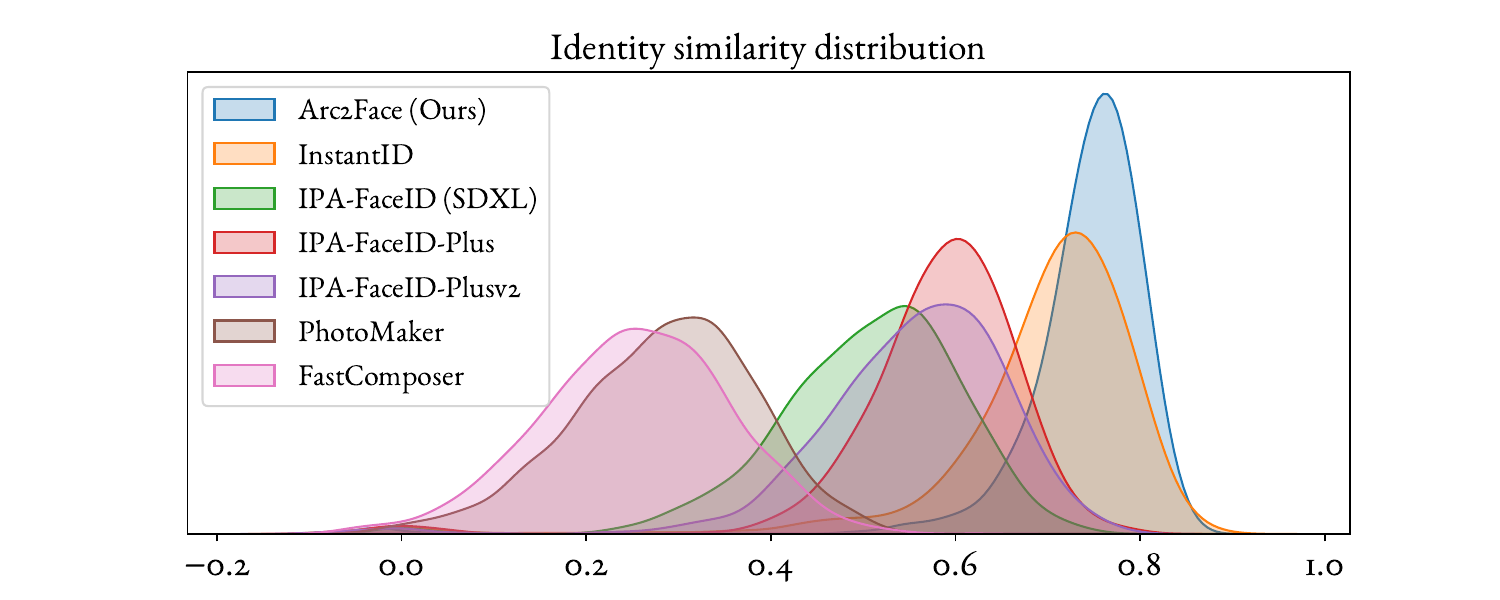}
      \caption{ID similarity \cite{kim2022adaface} between 400 input IDs and generated images of them (5 per ID) by different models.}
      \label{fig:id_sim_adaface}
  \end{minipage}
\end{figure}

Our model provides state-of-the-art ID-retention by relying solely on the ArcFace~\cite{deng2019arcface} embeddings of a single input image. However, in some cases, given multiple images of the same person, averaging the ID-embeddings may lead to slightly higher ID resemblance in the generation as shown in Fig.~\ref{fig:multiple_images}. Despite their highly discriminative nature, ID-embeddings from FR networks may still encode some ID-irrelevant features, such as expression or appearance, to a small extent. Consequently, averaging across diverse inputs proves beneficial in filtering out this noisy information.

\section{Retrieving the Closest Training Samples}
\begin{figure}
    \centering
    \includegraphics[width=0.9\textwidth]{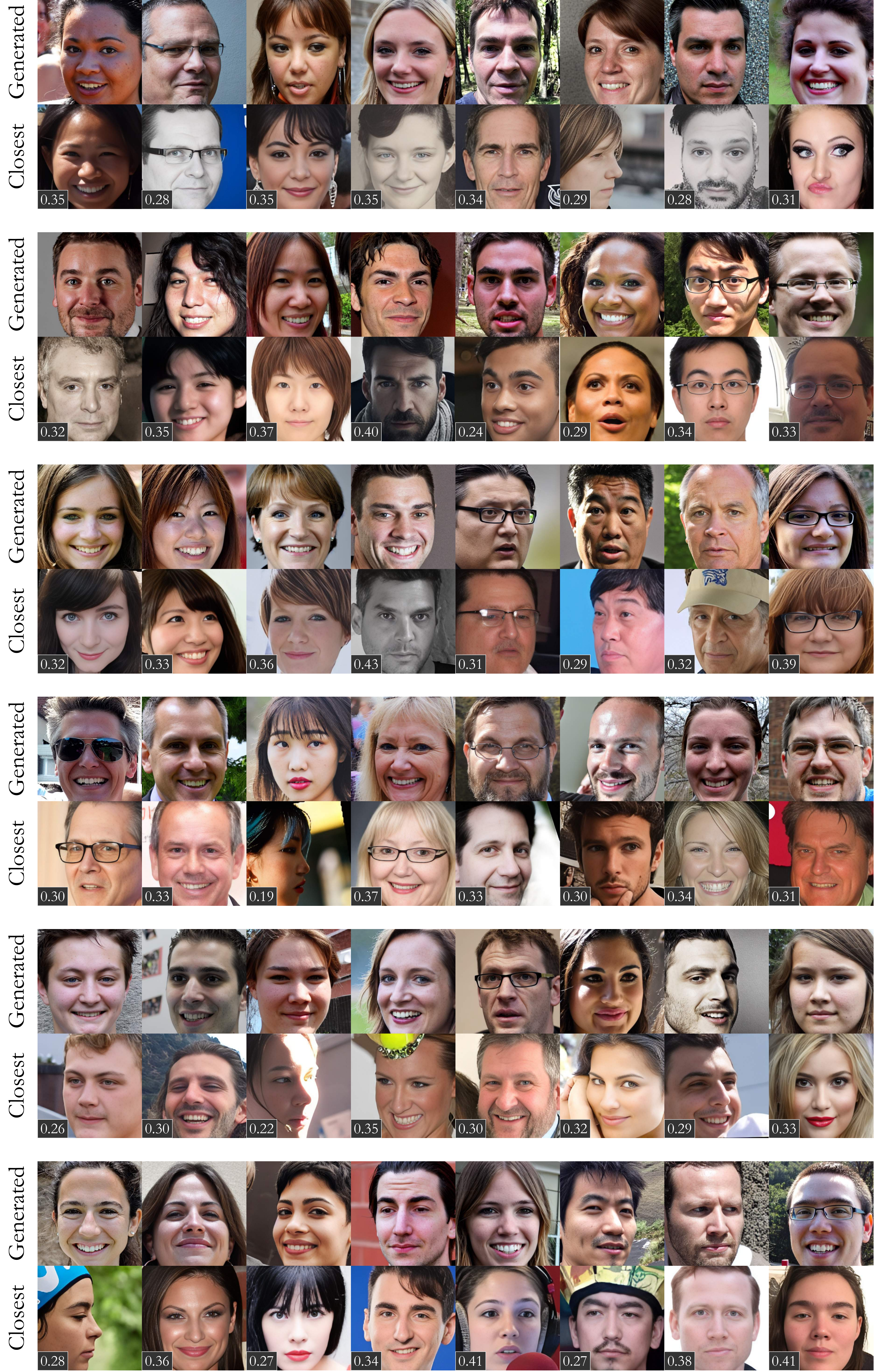}
    \caption{Faces generated from \modelname{} and their closest train samples, determined by ArcFace similarity (displayed at the bottom left). The generated faces correspond to ID vectors from our synthetic dataset (see Sec. \ref{sec:datasets}). For the purpose of comparison, the generated faces are cropped to the training face template.}
    \label{fig:closest_train_samples_ext}
\end{figure}

We have shown that, unlike the image memorization observed in T2I models~\cite{somepalli2023diffusion}, our model avoids replicating its training IDs, owing to our extensive training dataset. Fig. \ref{fig:closest_train_samples} shows examples of generated images from our model and their closest training images in terms of ID similarity. For a comprehensive analysis, we include additional examples in Fig.~\ref{fig:closest_train_samples_ext}, where a significant ID distance can be found between each sample and its closest match.

\section{Additional Qualitative Results}
\begin{figure}
    \centering
    \includegraphics[width=\textwidth, trim={0cm 1.5cm 0cm 1.5cm}, clip]{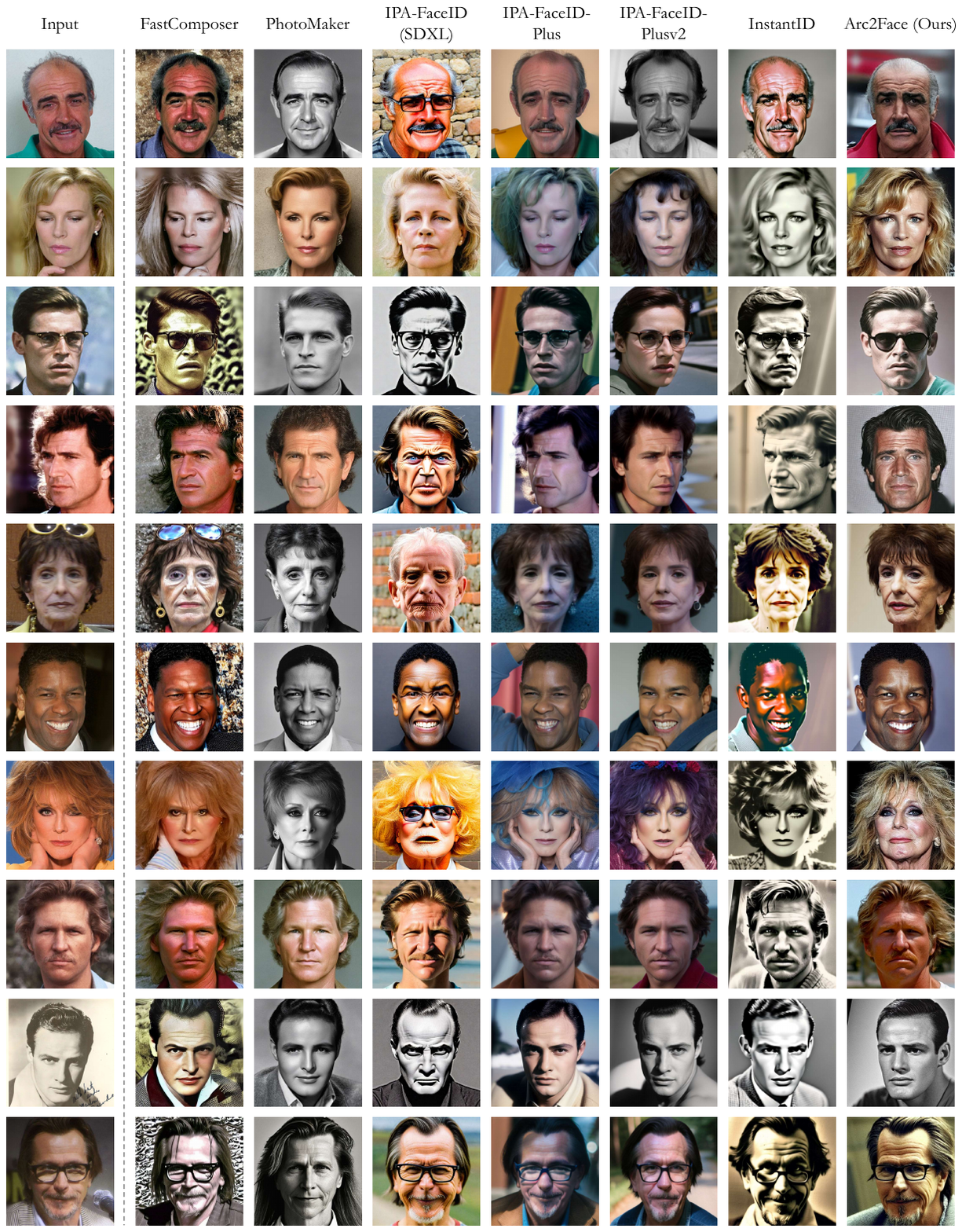}
    \caption{Comparison of \modelname{} with \cite{xiao2023fastcomposer, li2023photomaker, ye2023ip, wang2024instantid}. As described in the paper, we use the abstract prompt ``photo of a person'' for the text-based methods to focus on their ID-conditioning ability.}
    \label{fig:comp_ext}
\end{figure}

We expand our visual comparison with relevant methods~\cite{xiao2023fastcomposer, li2023photomaker, ye2023ip, wang2024instantid} in Fig.~\ref{fig:comp_ext}. Moreover, we compare their distributions of ID similarity with respect to the input in Fig.~\ref{fig:id_sim_adaface}. This comparison follows the evaluation protocol described in the main paper for our real-image dataset (refer Sec. \ref{sec:datasets}). However, we use a different FR network for ID-embedding extraction, namely AdaFace~\cite{kim2022adaface}, for completeness. The distributions closely resemble those presented in the main paper, confirming the superiority of \modelname{} in terms of ID retention. 

\begin{figure}
    \centering
    \includegraphics[width=\textwidth]{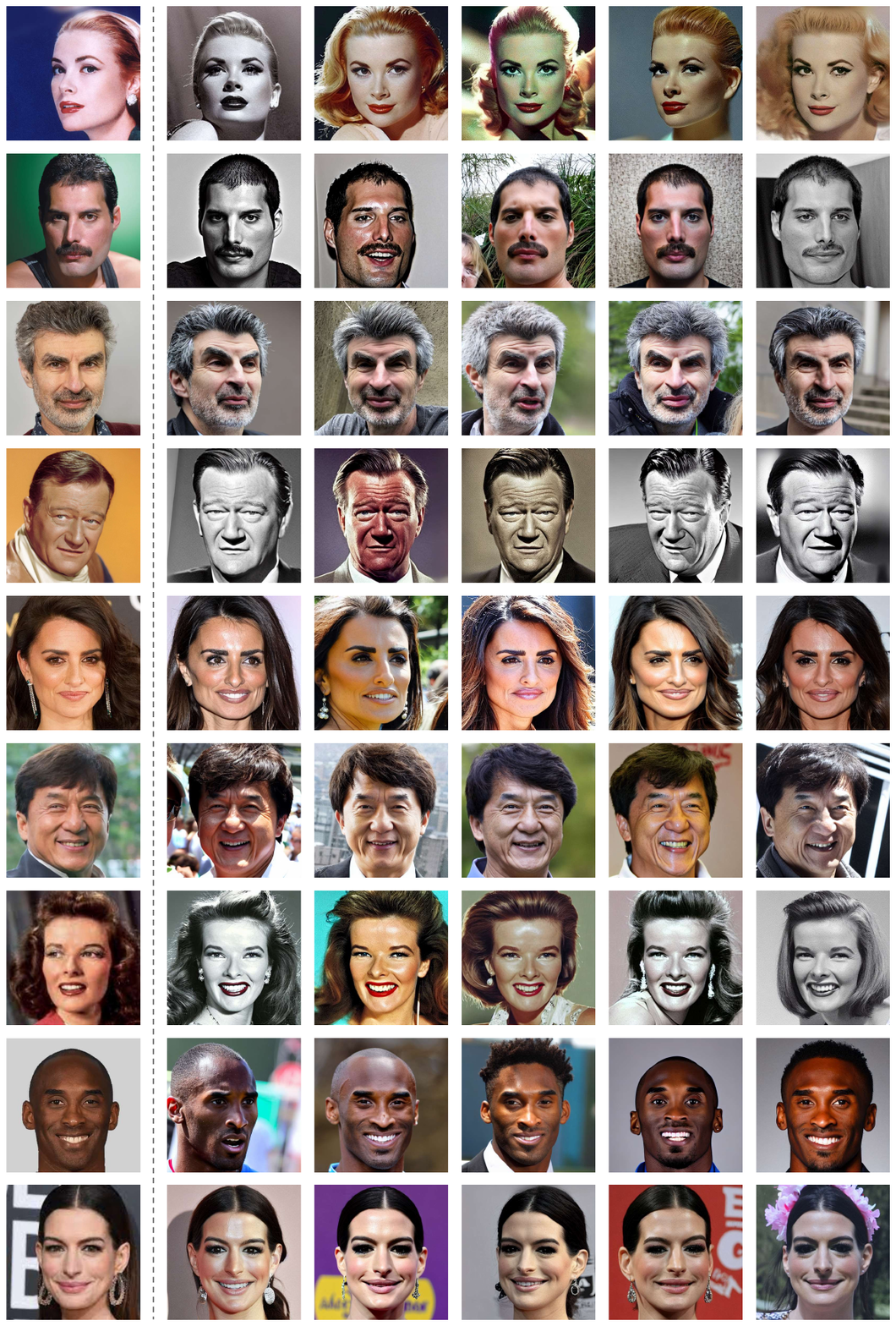}
    \caption{Multiple samples produced by our model conditioned on the input ID (leftmost column).}
    \label{fig:arc2face_samples1}
\end{figure}

\begin{figure}
    \centering
    \includegraphics[width=\textwidth]{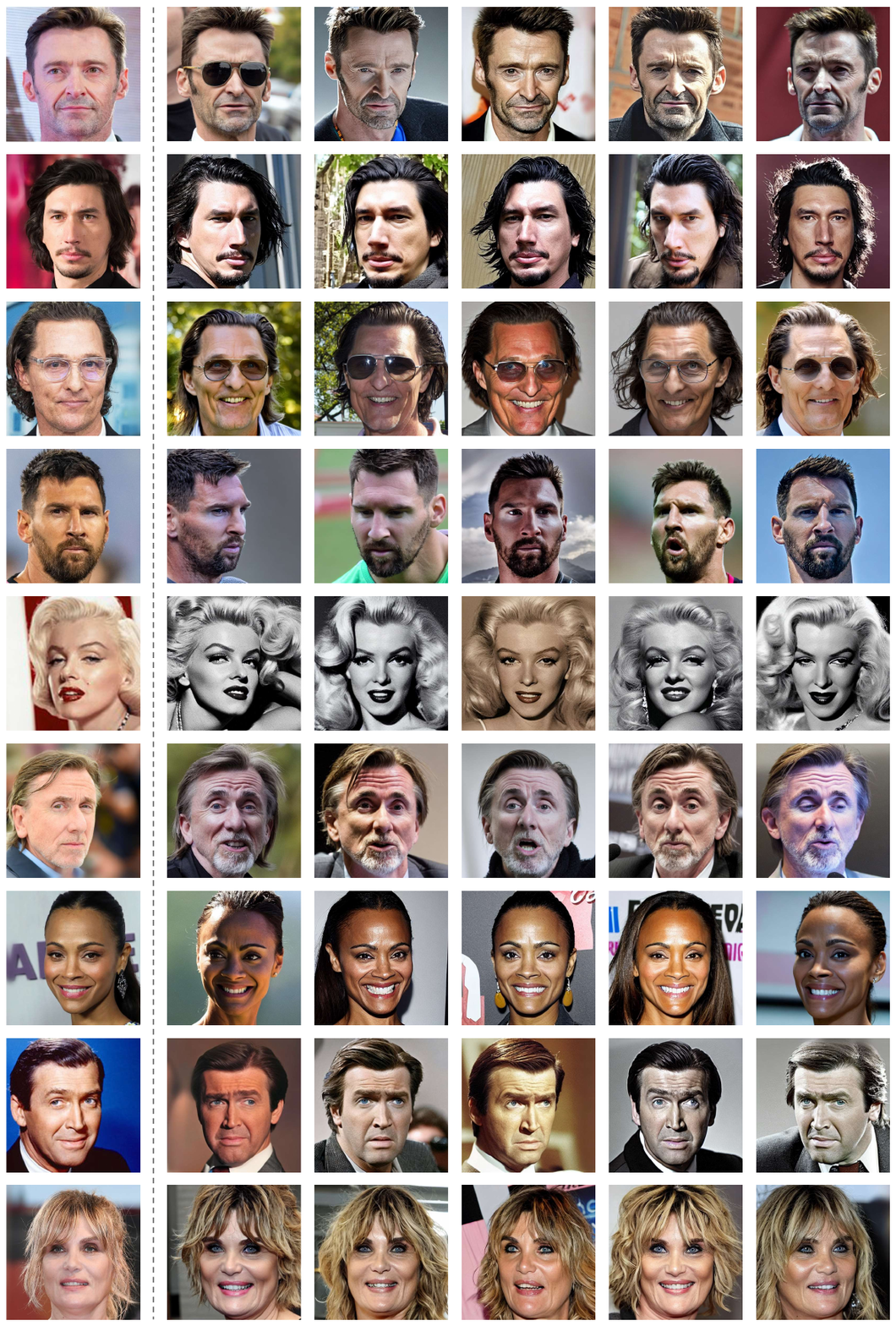}
    \caption{Multiple samples produced by our model conditioned on the input ID (leftmost column).}
    \label{fig:arc2face_samples2}
\end{figure}

\begin{figure}
    \centering
    \includegraphics[width=\textwidth]{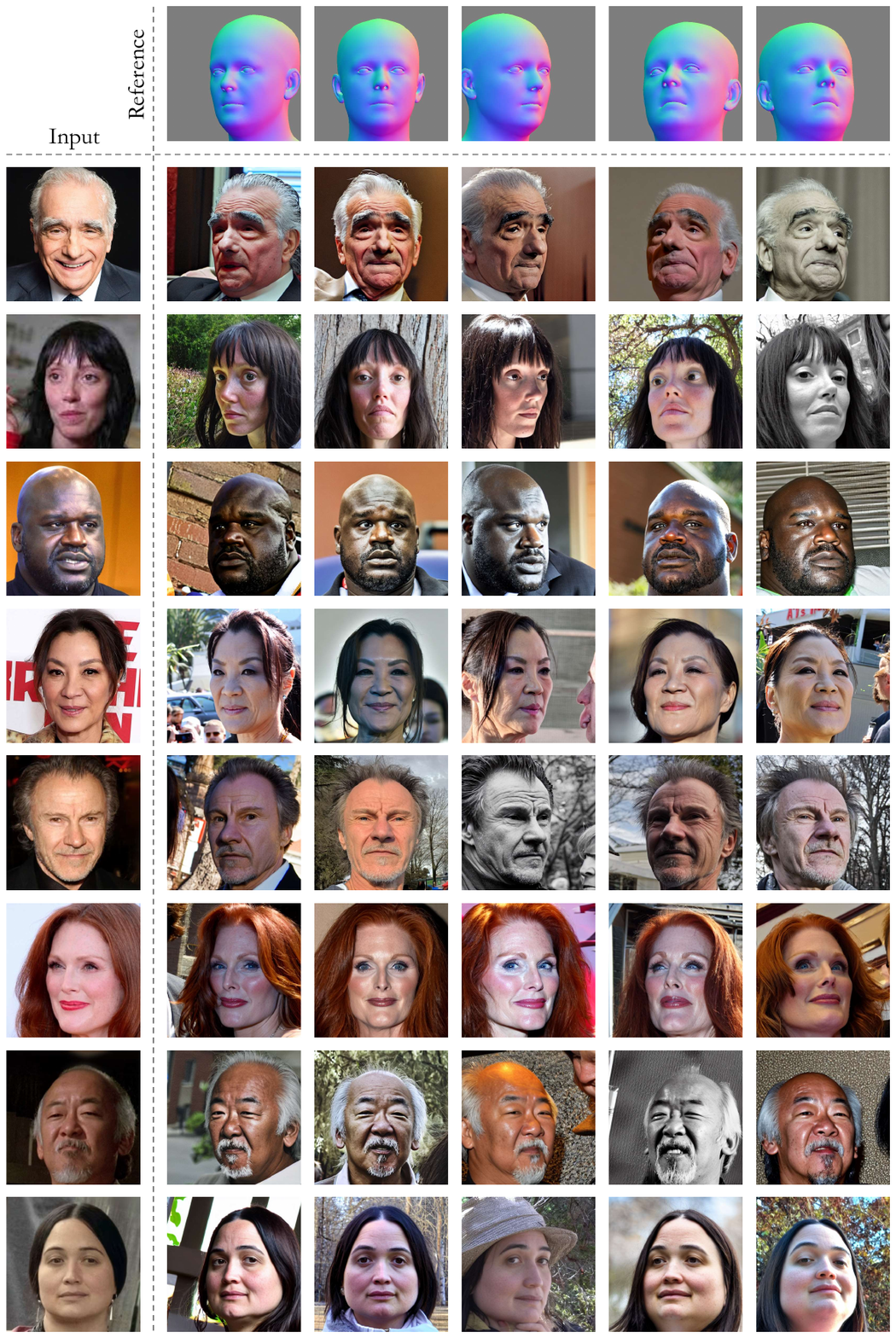}
    \caption{Additional results from \modelname{}, conditioned on a 3DMM~\cite{FLAME:SiggraphAsia2017, danvevcek2022emoca} using ControlNet~\cite{zhang2023adding}.}
    \label{fig:controlnet_ext_1}
\end{figure}

\begin{figure}
    \centering
    \includegraphics[width=\textwidth]{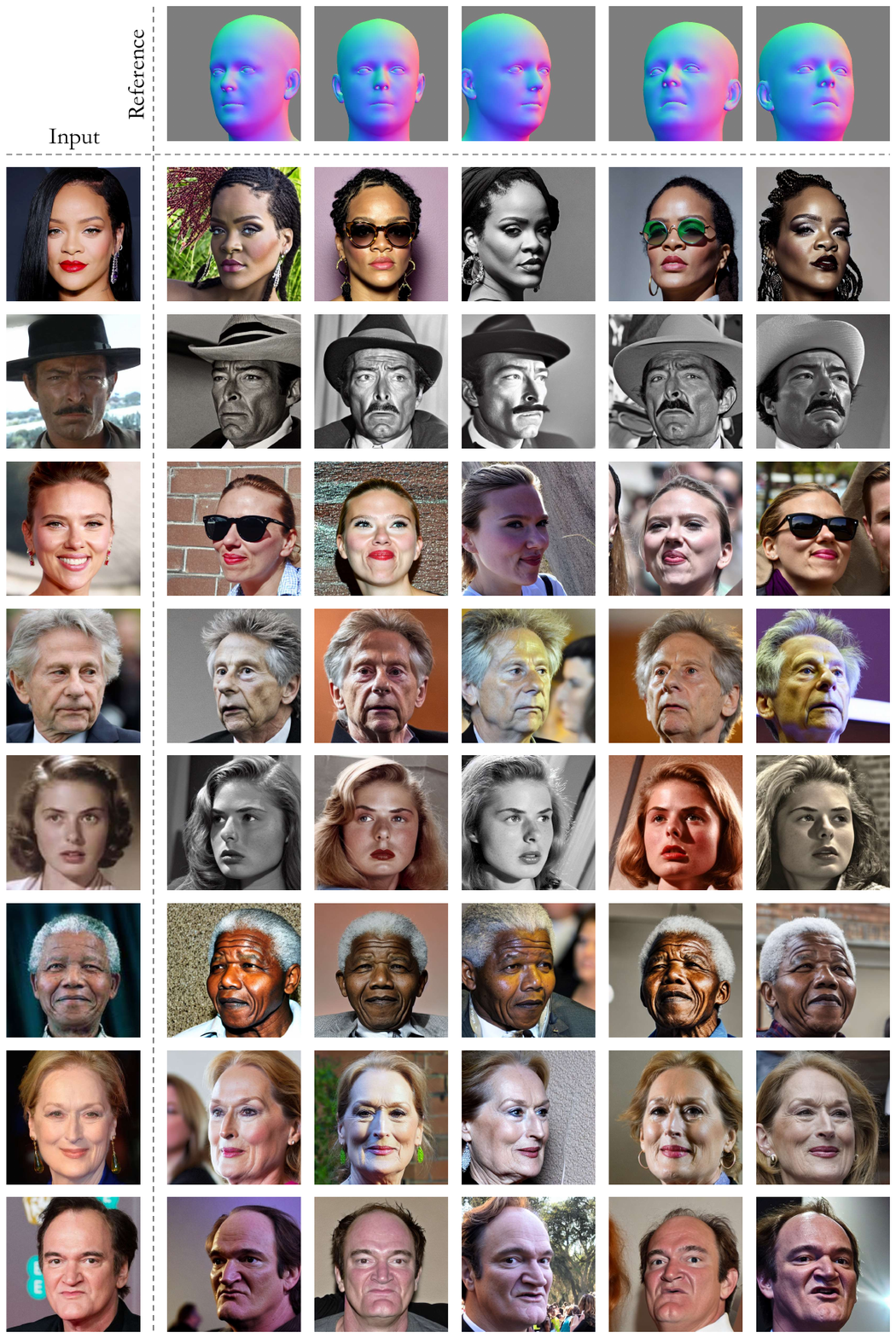}
    \caption{Additional results from \modelname{}, conditioned on a 3DMM~\cite{FLAME:SiggraphAsia2017, danvevcek2022emoca} using ControlNet~\cite{zhang2023adding}.}
    \label{fig:controlnet_ext_2}
\end{figure}

Finally, we provide additional visual results of our model in Fig.~\ref{fig:arc2face_samples1} and \ref{fig:arc2face_samples2}, as well as examples of our model combined with ControlNet~\cite{zhang2023adding} for explicit control in Fig. \ref{fig:controlnet_ext_1} and \ref{fig:controlnet_ext_2}.

\end{document}